\documentclass[]{bytedance_seed}

\usepackage{hyperref}
\usepackage{cleveref}
\usepackage{verbatim}

\usepackage{wrapfig}  
\usepackage{graphicx}
\usepackage{subcaption}
\usepackage{listings}
\usepackage{algorithm}
\usepackage{amsmath, amssymb}   
\usepackage{booktabs}           
\usepackage{multirow}           
\usepackage{makecell}           

\usepackage[table]{xcolor}
\usepackage{pifont}

\usepackage{subcaption} %

\usepackage[toc,page,header]{appendix}

\usepackage{minitoc}

\definecolor{rowblue}{HTML}{F0F6FF} 
\definecolor{rowgray}{HTML}{F5F5F5}
\definecolor{bestblue}{HTML}{E6F2FF} 
\newcommand{\cmark}{\textcolor{green!60!black}{\ding{51}}} 
\newcommand{\xmark}{\textcolor{red!60!black}{\ding{55}}} 
\newcommand{\tablestyle}[2]{\setlength{\tabcolsep}{#1}\renewcommand{\arraystretch}{#2}\centering\footnotesize}

\title{UniWeTok: An Unified Binary Tokenizer with Codebook Size $\mathit{2^{128}}$ for Unified Multimodal Large Language Model}

\author{%
\parbox{\textwidth}{\centering
Shaobin Zhuang$^{1,2,*}$,
Yuang Ai$^{1,3,*}$, 
Jiaming Han$^{1,3,*}$, 
Weijia Mao$^{1,5}$, 
Xiaohui Li$^{2}$\\[2mm]
Fangyikang Wang$^{6}$, 
Xiao Wang$^{1}$,
Yan Li$^{1}$,
Shanchuan Lin$^{1}$,
Kun Xu$^{1}$\\[2mm]
Zhenheng Yang$^{1}$,
Huaibo Huang$^{4,\dagger}$,
Xiangyu Yue$^{3}$, 
Hao Chen$^{1,*,\dagger,\ddagger}$,
Yali Wang$^{7,\dagger}$
}}

\affiliation{%
\parbox{\textwidth}{\centering\small
$^1$ByteDance,
$^2$Shanghai Jiao Tong University,
$^3$MMLab, The Chinese University of Hong Kong\\[1mm]
$^4$Institute of Automation, Chinese Academy of Sciences,
$^5$National University of Singapore\\[1mm]
$^6$Zhejiang University,
$^7$Shenzhen Institutes of Advanced Technology, Chinese Academy of Sciences
}}

\contribution[*]{Equal contribution}
\contribution[\dagger]{Corresponding Author}
\contribution[\ddagger]{Project lead}

\abstract{
\begin{abstract}

\label{sec:abs}

Unified Multimodal Large Language Models (MLLMs) require a visual representation that simultaneously supports high-fidelity reconstruction, 
complex semantic extraction, 
and generative suitability.
However, 
existing visual tokenizers typically struggle to satisfy these conflicting objectives within a single framework. 
In this paper, 
we introduce UniWeTok, 
a unified discrete tokenizer designed to bridge this gap using a massive binary codebook ($\mathit{2^{128}}$).
For training framework,
we introduce Pre-Post Distillation and a Generative-Aware Prior to enhance the semantic extraction and generative prior of the discrete tokens. 
In terms of model architecture,
we propose a convolution-attention hybrid architecture with the SigLu activation function.
SigLu activation not only bounds the encoder output and stabilizes the semantic distillation process but also effectively addresses the optimization conflict between token entropy loss and commitment loss.
We further propose a three-stage training framework designed to enhance UniWeTok’s adaptability across various image resolutions and perception-sensitive scenarios, 
such as those involving human faces and textual content.
On ImageNet, 
UniWeTok achieves state-of-the-art image generation performance (FID: UniWeTok 1.38 \textit{vs.} REPA 1.42) while requiring a remarkably low training compute (Training Tokens: UniWeTok 33B \textit{vs.} REPA 262B). On general-domain, 
UniWeTok demonstrates highly competitive capabilities across a broad range of tasks, including multimodal understanding, 
image generation (DPG Score: UniWeTok 86.63 \textit{vs.} FLUX.1 [Dev] 83.84),
and editing (GEdit Overall Score: UniWeTok 5.09 \textit{vs.} OmniGen 5.06).
We release code and models to facilitate community exploration of unified tokenizer and MLLM.
  
\end{abstract}
}

\checkdata[Code]{\url{https://github.com/shallowdream204/BitDance}}

\begin{document}
\maketitle

\begin{figure*}
    \centering
    \makebox[\linewidth][c]{%
        \begin{minipage}{1.3\linewidth}
            \centering
            \includegraphics[width=\linewidth]{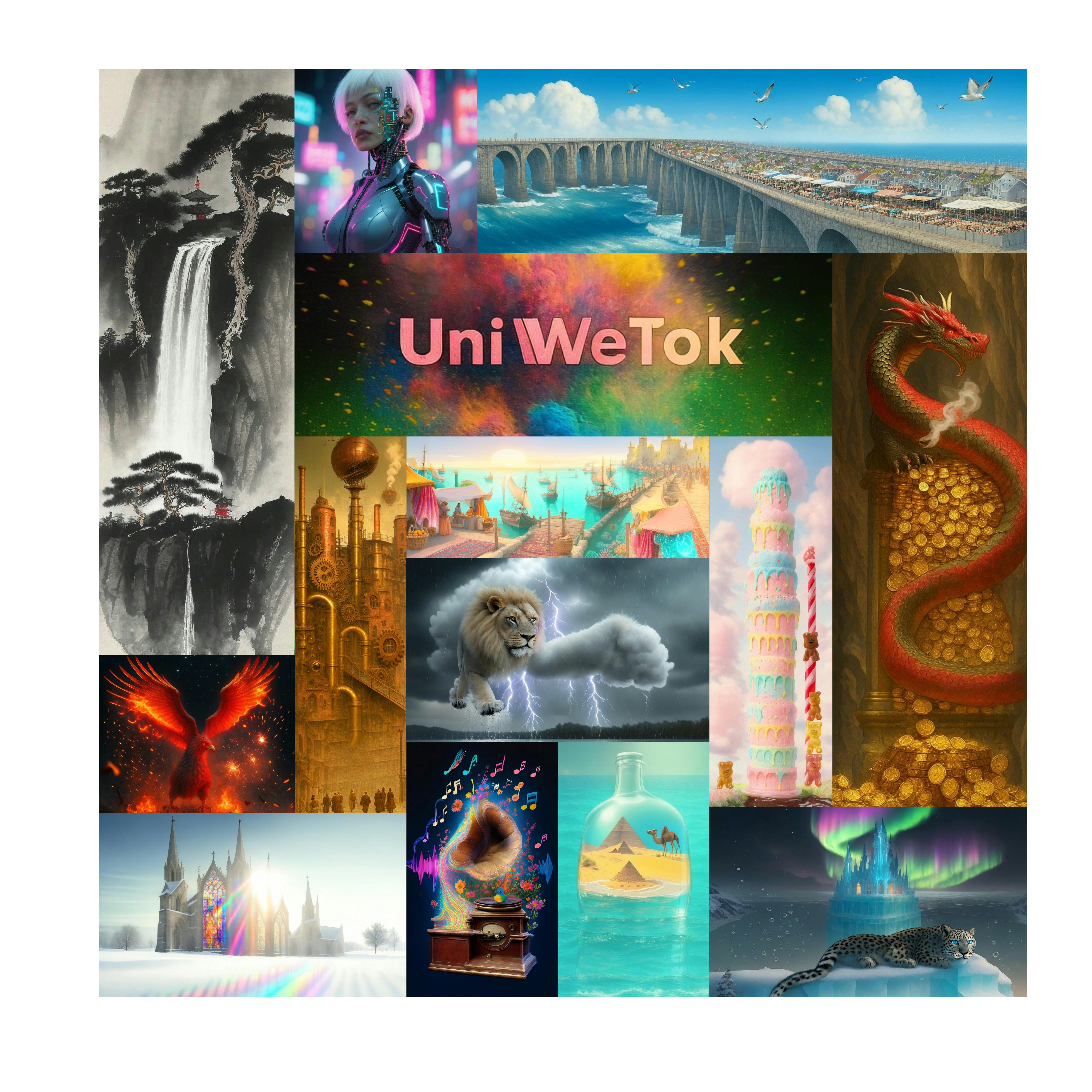}
            \caption{\textbf{Various resolution samples generated by our Unified MLLM based on UniWeTok}, 
showcasing its capabilities in prompt adherence, spatial reasoning, and text rendering across various artistic styles.
UniWeTok employs a spatial downsampling rate of 32$\times$ and a codebook size of $2^{128}$, 
which reduces the number of visual tokens by 75\% while maintaining exceptionally high reconstruction quality. 
Furthermore, 
UniWeTok possesses strong semantic extraction capabilities and generative priors, 
making compressed discrete visual tokens suitable for Unified Multimodal Large Language Model.
            }
            \label{fig:teaser}
        \end{minipage}%
    }
\end{figure*}

\section{Introduction}

The success of the next-token prediction paradigm in Large Language Models (LLMs)~\citep{radford2018improving,radford2019language,mann2020language,achiam2023gpt} has motivated extensive research into transferring this approach to unified vision-language modeling.
However, 
achieving this unification presents significant challenges.
A primary obstacle is the prohibitive computational cost associated with pixel-level modeling, 
which necessitates effective image compression.
Visual tokenizers address this by employing an encoder-decoder architecture~\citep{Kingma2013AutoEncodingVB} to compress images into compact latent representations for subsequent reconstruction.
While employing continuous tokenizers to extract image latents is a common approach, 
it frequently suffers from error accumulation and mode collapse during autoregressive generation~\citep{team2025nextstep,zhuang2025wetok}. 
Consequently, 
modeling the distribution of discrete tokens offers a more robust alternative.

Despite their robustness, 
discrete tokenizers have historically faced criticism for limited reconstruction capabilities, implying significant information loss.
Recent advancements have fundamentally shifted this landscape. 
Through improvements to LFQ~\citep{yulanguage}, 
BSQ~\citep{zhao2024image} and GQ~\citep{zhuang2025wetok} scaled the codebook size to an unprecedented size—exceeding $2^{128}$—thereby enabling individual tokens to encapsulate a massive amount of information.
However, 
such an expansive codebook size introduces new complexities for downstream generation and understanding tasks. 
While recent approaches such as Infinity~\citep{han2025infinity} and BitDance~\citep{bitdance} propose solutions leveraging massive codebooks, 
the codebook size in text-to-image modeling remains constrained to $2^{32}$.
Furthermore, 
these approaches do not extend to multimodal understanding or the development of Unified Multimodal Large Language Models (MLLMs). 
This naturally leads to a question: \textit{Is it feasible to construct a Unified MLLM based on a massive discrete visual codebook?}

To achieve this, 
we propose UniWeTok, 
a visual discrete tokenizer that unifies robust compression, 
semantic extraction, 
and generative priors into a single framework. 
UniWeTok maximizes token information density by achieving $32\times$ spatial downsampling while maintaining a codebook size of $2^{128}$. 
Building upon WeTok~\citep{zhuang2025wetok}, 
we comprehensively advance the system across three key dimensions: (1) Training Framework: 
We introduce a Pre-Post Distillation (PPD) loss and a Generative-Aware Prior (GAP) loss to significantly enhance performance in downstream understanding and generation tasks.
(2) Model Architecture: 
We propose a SigLu activation to ensure the stable convergence of the PPD loss. 
Furthermore,
we transition to a hybrid backbone that synergizes the local inductive priors of convolutions with the global receptive field of attention mechanisms. 
(3) Training Pipeline: 
We introduce a three-stage curriculum learning strategy. 
By adjusting resolutions and training data distributions, 
we enable our UniWeTok to robustly adapt to variable resolutions and perceptually sensitive scenarios, such as human faces and text.

We first demonstrate that UniWeTok achieves state-of-the-art generation performance 
(FID: \textbf{UniWeTok 1.38 \textit{vs.} REPA} \textbf{1.42}) 
in class-to-image generation while incurring significantly lower training costs compared to existing methods 
(Training Tokens: \textbf{UniWeTok 33B \textit{vs.} REPA 262B}). 
Building on this efficiency, 
our Unified MLLM based on UniWeTok not only exhibits competitive capabilities in multimodal understanding but also delivers text-to-image generation quality that surpasses the prominent open-source model 
(DPG Score: \textbf{UniWeTok 86.63 \textit{vs.} FLUX.1 [Dev]} \textbf{83.84}). 
Furthermore,
our Unified MLLM also demonstrates remarkable proficiency in image editing 
(GEdit Overall Score: \textbf{UniWeTok 5.09 \textit{vs.} OmniGen} \textbf{5.06}). 
Collectively, 
these results rigorously validate the effectiveness of UniWeTok as the visual tokenizer for Unified MLLMs modeling.
\section{Related Work}

\subsection{Discrete Visual Tokenizer}

VQVAE \citep{van2017neural} and VQGAN \citep{esser2021taming} employ vector-quantization (VQ) to transform visual input into discrete tokens.
But they suffer from low reconstruction quality caused by instability of the codebook utilization.
To overcome these drawbacks, 
one line of work introduces optimization strategies or modules to improve performance \citep{lee2022autoregressive, shi2024scalable, zhu2024scaling, yu2024image}.
Another line of work focuses on scaling up the codebook size by grouping codebooks \textcolor{blue}{\citep{Ma2025UniTokAU, jia2025mgvq, zhang2025quantize,Bai2024FactorizedVT}}. 
ImageFolder \citep{Li2024ImageFolderAI}, 
DualToken \citep{Song2025DualTokenTU} and TokenFlow \citep{Qu2024TokenFlowUI} use multiple codebooks to assist in optimizing model understanding and reconstruction capabilities.
However, 
VQ-based tokenizers still introduce additional costs due to the lookup operation \citep{yu2021vector, lee2022autoregressive, fang2025enhancing}.
MAGVIT-v2 \citep{yu2024languagemodelbeatsdiffusion} introduces Lookup-Free Quantization to address extra cost and proposes the entropy loss \citep{Chang2022MaskGITMG, Jansen2019CoincidenceCA} to ensure the utilization of the codebook.
BSQ \citep{Zhao2024ImageAV} assumes
independence between the bits of the binary code to eliminate unbearable computational overhead from entropy loss.
WeTok~\citep{wetok} proposes Group-Wise Lookup-Free Quantization to mitigate the codebook entropy calculation error in BSQ.
However,
current tokenizers based on binary codebooks barely extract any semantic information~\citep{Qu2024TokenFlowUI}, 
and their excessively large codebooks are detrimental to downstream generation tasks~\citep{wetok}. 
In contrast, 
our UniWeTok successfully achieves robust semantic extraction performance based on a binary codebook while ensuring that the extracted discrete tokens remain suitable for downstream generation.

\subsection{Unified Multimodal Large Language Model}
With the development of LLMs~\cite{touvron2023llama,bai2023qwen,mann2020language,achiam2023gpt}, 
MLLMs have attracted a lot of research interest due to their strong multimodal understanding and reasoning capabilities~\cite{llava,li2023blip,dai2023instructblip,bai2023qwen}. 
Beyond visual understanding, 
several recent works~\cite{ge2023making,dong2023dreamllm,sun2024generative,showo,zhou2024transfusion,wang2025illume,wu2025janus} attempt to integrate both visual understanding and generation within a unified MLLM. 
Emu2~\cite{sun2024generative} enables LLMs to generate CLIP embeddings, 
which are decoded into images using a diffusion model. 
Show-o~\cite{showo} and Transfusion~\cite{zhou2024transfusion} integrate diffusion objectives into LLMs for image generation, 
but this design breaks the autoregressive paradigm and complicates the unification of the two tasks.
Emu3~\cite{emu3} and Chameleon~\cite{chameleon} use VQVAE~\cite{vqgan} as both the visual encoder and decoder, 
allowing unified next token prediction across images and text.
Janus~\cite{wu2025janus,chen2025janus} employs separate encoders for understanding and generation, 
resulting in distinct modalities that limit performance in multi-turn editing and interleaved generation.
VILA-U~\cite{wu2024vila} and UniTok~\cite{unitok} are trained using both pixel reconstruction and image-text alignment losses, 
but they struggle to converge optimally for both tasks.  
In contrast, 
we propose a discrete, semantic and generation-friendly that unifies understanding and generation within a single MLLM.

\section{Method}
\label{sec:method}

\begin{wrapfigure}{l}{0.42\textwidth}
    \includegraphics[width=\linewidth]{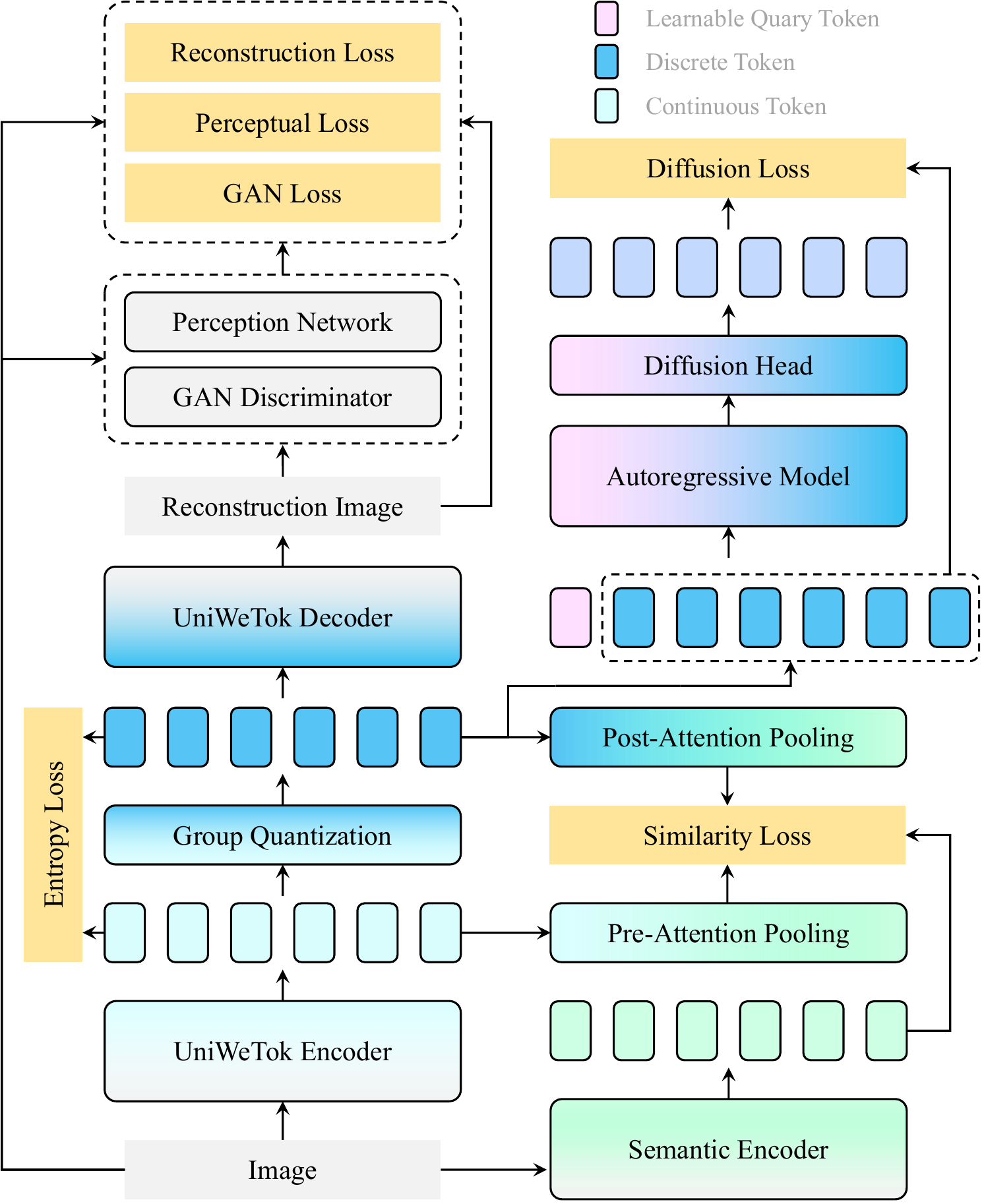}
    \caption{\textbf{Illustration of UniWeTok training framework.}  
    We introduce a pre-trained semantic encoder for Pre-Post Distillation and a lightweight generative model for Generative-Aware Prior.
    }
    \label{fig:train_framework}
\end{wrapfigure}

In this section, 
we present the proposed UniWeTok methodology. 
We begin by introducing the Pre-Post Distillation (PPD) and Generative-Aware Prior (GAP) in Sec.~\ref{subsec:train_framework}, 
which serve as the foundational elements of our training framework. 
Subsequently, 
we elaborate on the specific structural design in Sec.~\ref{subsec:model_arch},
focusing on the SigLu activation and the hybrid architecture. 
Finally, 
we demonstrate our three-stage curriculum training pipeline in Sec.~\ref{subsec:train_pipe},
a critical mechanism that enables UniWeTok to generalize robustly across variable resolutions and high-sensitivity scenarios.

\subsection{Training Framework
}
\label{subsec:train_framework}

\textbf{Preliminaries.}
WeTok~\citep{zhuang2025wetok} compress image $\mathcal{I} \in \mathbb{R}^{H \times W \times 3}$
into latent feature $\mathcal{U}$$=$$\mathcal{E}(\mathcal{I})$,
$\mathcal{U}$$\in$$\mathbb{R}^{h \times w \times d}$,
through the encoder $\mathcal{E}$.
It groups the latent features in channel dimension, 
reshape $\mathcal{U}$ into $\mathcal{U}_{\text{G}}$$\in$$\mathbb{R}^{h \times w \times g \times d'}$,
where $d$$=$$gd'$ and $g$ and $d'$ represent the number and
channel of groups.
Group-Wise Lookup-Free Quantization (GQ) introduces codebooks $\mathcal{C}_{\text{GQ},k}$$=$$\{ -1, 1 \}^{d'}$ to perform lookup-free quantization on each channel of latent feature,
\begin{equation}
\mathcal{Q}[i,j,k,l] = \text{sign}(\mathcal{U}_{\text{G}}[i,j,k,l]).
\label{eq:gq_quantize}
\end{equation}
Notice that we introduce $\mathcal{U}_{\mathcal{Q}} = \mathcal{U}_{\text{G}}+\text{sg}[\mathcal{Q}-\mathcal{U}_{\text{G}}]$ as a variable that shares the same value as $\mathcal{Q}$ but has a backward gradient only with respect to $\mathcal{U}_{\text{G}}$. Here, $\text{sg}[\cdot]$ denotes the stop-gradient operation.
The $\mathcal{U}_{\mathcal{Q}}$ is reconstructed into image space $\hat{\mathcal{I}} = \mathcal{G}(\mathcal{U}_{\mathcal{Q}})$ through the decoder $\mathcal{G}$. 
The loss function of WeTok consists of the following five parts,
\begin{align}
\mathcal{L}_{\text{WeTok}} &= \underbrace{\| \mathcal{I} - \hat{\mathcal{I}} \|^{2}}_{\text{Recon. Loss}} + \alpha \underbrace{\| \mathcal{U}_{\text{G}} - \text{sg}[\mathcal{Q}] \|^{2}}_{\text{Commitment Loss}} + \beta \underbrace{\mathcal{L}_{\text{LPIPS}}(\mathcal{I}, \hat{\mathcal{I}})}_{\text{Perceptual Loss}} + \gamma \underbrace{\mathcal{L}_{\text{GAN}}(\mathcal{I}, \hat{\mathcal{I}})}_{\text{GAN Loss}} + \delta \underbrace{\mathcal{L}_{\text{Entropy}}(\mathcal{U}_{\text{G}},\mathcal{Q})}_{\text{Entropy Loss}}, 
\label{eq:wetok_loss}
\end{align}
where perceptual loss \citep{Zhang2018TheUE} and GAN loss are introduced for better visual quality.
Entropy loss consists of token entropy loss and codebook entropy loss. The specific form of token entropy loss is as follows:
\begin{equation}
\mathcal{L}_{\text{Token Entropy}}
=\frac{1}{hw} \sum_{i=1}^{h} \sum_{j=1}^{w} \sum_{k=1}^{g} H(q_{\text{G}}(\mathbf{c}_{k} | \mathcal{U_{\text{G}}}[i,j,k])),
\label{eq:gfq_token_loss}
\end{equation}
where $q_{\text{G}}(\mathbf{c}_{k} | \mathcal{U_{\text{G}}}[i,j,k])$ denote the conditional distribution of $\mathbf{c}_k$$\in$$\mathcal{C}_{\text{GQ},k}$ given
$\mathcal{U_{\text{G}}}[i,j,k]$.

\textbf{Pre-Post Distillation.}
To facilitate the effective application of UniWeTok in multimodal understanding tasks, 
it is imperative to endow the encoder with semantic extraction capabilities. 
As illustrated in Fig.~\ref{fig:train_framework}, 
we employ a pre-trained semantic encoder $E_{\text{T}}$ as the teacher for distillation. 
Specifically, 
the teacher encodes the input image $\mathcal{I}$ into semantic latents $f_t$$=$$E_{\text{T}}(\mathcal{I})$,
$f_t$$\in$$\mathbb{R}^{1 \times d_{\text{T}}}$,
where $d_{\text{T}}$ is the output dimension of $E_{\text{T}}$.
We employ cosine similarity loss for distillation. Notably, 
we align both $\mathcal{U}_{\text{G}}$ and $\mathcal{U}_{\mathcal{Q}}$ with $f_{\text{T}}$,
\begin{equation}
\mathcal{L}_{\text{Pre Distill}}
=1-\frac{u_{\text{G}} \cdot f_{\text{T}}}{||u_{\text{G}}||||f_{\text{T}}||},
u_{\text{G}}=\text{AttnPool}_{\text{Pre}}(\mathcal{U}_{\text{G}}),
\label{eq:pre_dis}
\end{equation}
\begin{equation}
\mathcal{L}_{\text{Post Distill}}
=1-\frac{u_{\mathcal{Q}} \cdot f_{\text{T}}}{||u_{\mathcal{Q}}||||f_{\text{T}}||},
u_{\mathcal{Q}}=\text{AttnPool}_{\text{Post}}(\mathcal{U}_{\mathcal{Q}}),
\label{eq:post_dis}
\end{equation}
\begin{equation}
\mathcal{L}_{\text{PPD}}
=\mathcal{L}_{\text{Pre Distill}} + \eta\mathcal{L}_{\text{Post Distill}}.
\label{eq:ppd_loss}
\end{equation}
This alignment strategy serves to enhance the semantic extraction capability, 
ensuring that the model effectively captures the underlying semantic information.

\begin{figure*}
    \centering
    \includegraphics[width=\linewidth]{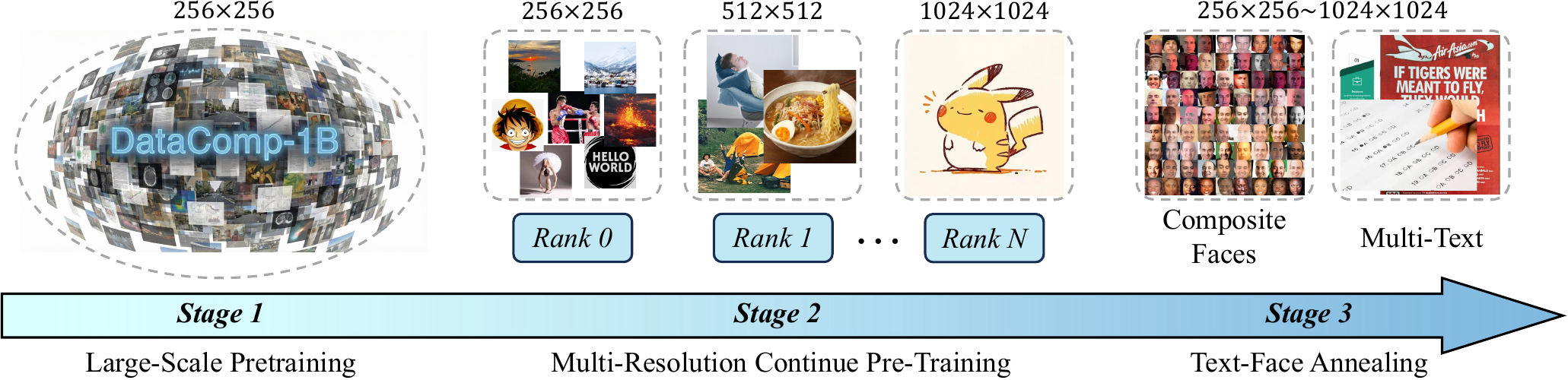}
    \caption{\textbf{Illustration of the three-stage training pipeline of UniWeTok.}
    }
    \label{fig:train_pipe}
\end{figure*}
\begin{figure}[!ht]
    \centering
    \includegraphics[width=\linewidth]{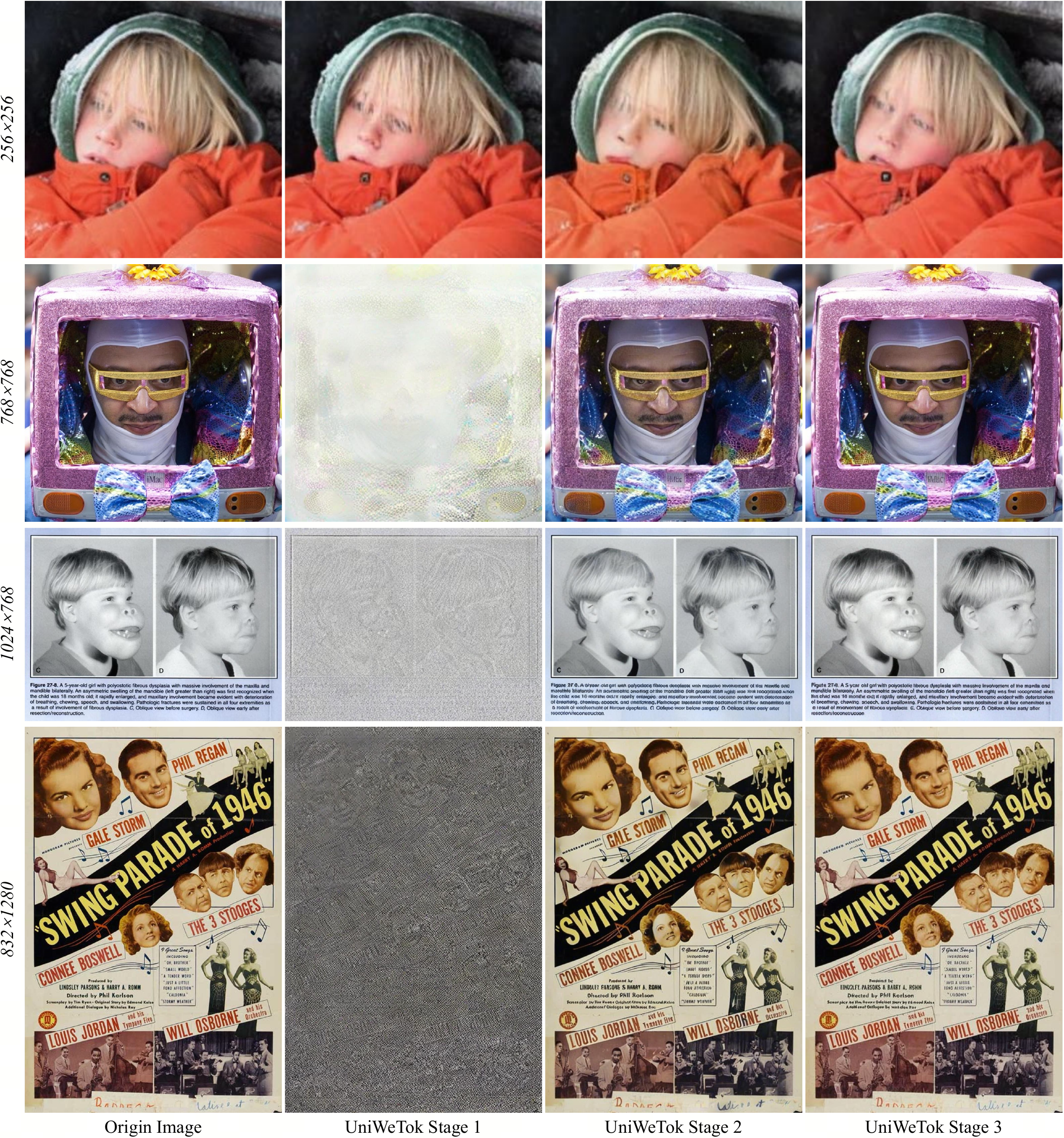}
    \caption{\textbf{Ablation study of three-stage training pipeline.}
    }
    \label{fig:train_pipe_ablation}
\end{figure}

\textbf{Generative-Aware Prior.}
Increasing the number of groups significantly enhances the reconstruction performance of the discrete tokenizer~\citep{zhuang2025wetok}.
However, 
this improvement comes at the cost of increased difficulty for downstream generation tasks. 
To mitigate this issue, 
we inject a Generative-Aware Prior, 
enabling the model to perceive the generation objective during the training phase. 
As illustrated in Fig.~\ref{fig:train_framework}, 
we flatten the $\mathcal{U}_{\mathcal{Q}}$ into 1D sequence $\mathcal{U}_{\mathcal{B}}$$\in$$\mathbb{R}^{(h \times w) \times d}$ and feed it into a randomly initialized tiny BitDance~\citep{bitdance} model $B$ to perform a next-token diffusion task, 
which is supervised using the MSE loss,
\begin{equation}
\mathcal{L}_{\text{GAP}}
=\mathbb{E}_{\epsilon\in \mathcal{N}\text{(\textbf{0},\textbf{I})}}||\mathcal{U}_{\mathcal{B}}  - B([\text{Query},\mathcal{U}_{\mathcal{B}}[:-1]],\epsilon,t)||^2,
\label{eq:gap_loss}
\end{equation}
where $t$ is random sampled from $[0,1]$.
Query token allows the model not to miss the prediction of the first token and learn the distribution of the entire sequence.

Consequently, the comprehensive loss function for the UniWeTok framework is defined as follows:
\begin{equation}
\mathcal{L}_{\text{UniWeTok}}
=\mathcal{L}_{\text{WeTok}} + \theta \mathcal{L}_{\text{PPD}} + \mu \mathcal{L}_{\text{GAP}},
\label{eq:uniwetok_loss}
\end{equation}
where $\theta$ and $\mu$ serve as hyperparameters that to regulate the weight of reconstruction, 
understanding and generation task during the UniWeTok training process.

\subsection{Unified Discrete Tokenizer Architecture}
\label{subsec:model_arch}

\textbf{Hybrid backbone.}
UniWeTok employs a hybrid architecture for both encoder and decoder. 
In the encoder, 
input images are initially processed by stacked residual blocks, 
which serve to extract low-level information and perform spatial downsampling. 
This is followed by a sequence of transformer blocks designed to capture global context. 
The decoder mirrors this structure symmetrically. 
As shown in Tab.~\ref{tab:hybrid},
this design effectively integrates the local inductive bias of convolutional layers while minimizing the computational overhead associated with the attention mechanism.
Notably, 
the original WeTok architectu utilizes a downsample block that executes spatial downsampling before channel expansion,
a sequential process that results in information loss. 
To address this,
we modify the downsample block to perform channel expansion concurrently with downsampling. 
As shown in Tab.~\ref{tab:channel}, 
this modification yields a substantial improvement in the reconstruction capabilities.

\textbf{SigLu activation.}
During the training process,
we observe a notable phenomenon. 
As shown in Tab.~\ref{tab:distill_strat},
when we perform semantic distillation exclusively on $\mathcal{U}_{\mathcal{Q}}$,
the model fails to capture meaningful semantic information. 
We attribute this issue to the commitment loss, 
which rigidly anchors the encoder outputs $\mathcal{U}_{\text{G}}$ to values near -1 or 1, 
making it difficult for the model to adapt its semantic representations.
This constraint also creates a fundamental conflict with the token entropy loss, as the latter drives the $\mathcal{U}_{\text{G}}$ towards negative or positive infinity. 
To resolve this optimization conflict, 
we propose the SigLu activation function,
\begin{equation}
\text{SigLu}(x)=\frac{1-e^{x}}{1+e^{x}},
\label{eq:siglu}
\end{equation}
where SigLu is integrated as the final layer of the encoder.
The SigLu activation inherently constrains $\mathcal{U}_{\text{G}}$ to the interval $[-1, 1]$. 
Under this condition,
the token entropy loss becomes equivalent to the commitment loss. 
Consequently, 
we set $\alpha$$=$$0$ in Eq.~\ref{eq:wetok_loss},
effectively replacing the commitment term with the token entropy loss. 
As shown in Tab.~\ref{tab:distill_strat},
SigLu enables the model to perform semantic distillation stably.
The specific model architecture could be seen in App.~\ref{app:model_arch}.

\subsection{Training Pipeline}
\label{subsec:train_pipe}

As illustrated in Fig. \ref{fig:train_pipe_ablation}, 
our UniWeTok model is constrained to the specific image resolution defined during its pre-training. 
However, practical downstream applications inevitably require handling a diverse range of resolutions. 
To bridge this gap, we propose a three-stage progressive pre-training strategy. 
As depicted in Fig. \ref{fig:train_pipe}, 
the first stage prioritizes computational efficiency by performing large-scale pre-training on a general-domain dataset at a fixed resolution of $256$$\times$$256$. 
In the second stage, 
we partition computational resources to facilitate training across multiple resolutions simultaneously.
Finally, 
the third stage employs an annealing training phase focused on perceptually sensitive domains, such as faces and text. 
This progressive refinement ensures that UniWeTok is optimally aligned with complex downstream understanding and generation tasks.

\section{Experiments}
\label{sec:experiments}

\textbf{Datasets.}
We train on two datasets: (i) ImageNet~\citep{Russakovsky2014ImageNetLS} training set; 
and 
(ii)
the general-domain dataset DataComp-1B~\citep{gadre2023datacomp}. 
For evaluations on ImageNet, we measure the reconstruction (rFID~\citep{Heusel2017GANsTB}, 
PSNR, 
SSIM~\citep{Wang2004ImageQA}, 
and LPIPS~\citep{Zhang2018TheUE})
and semantic extraction 
(zero-shot classification~\citep{radford2021learning})
performance on the ImageNet 50k validation set 
as well as the generation performance of the generative model (gFID, IS, Precision, and Recall). 
Regarding general-domain, 
we evaluate the reconstruction performance on the ImageNet 50k validation set and the MS-COCO 2017 validation set \citep{Lin2014MicrosoftCC}. To comprehensively assess the capabilities of UniWeTok in downstream scenarios, we train a Unified MLLM based on UniWeTok and evaluate its performance on downstream understanding 
(SEEDB~\citep{li2023seed},
POPE~\citep{li2023evaluating}, VQAv2~\citep{goyal2017making},
GQA~\citep{hudson2019gqa}, 
SQA~\citep{lu2022learn},
TQA~\citep{singh2019towards},
CQA~\citep{masry2022chartqa},
AI2D~\citep{kembhavi2016diagram},
RWQA~\citep{rwqa},
MMMU~\citep{yue2024mmmu}, 
MME~\citep{fu2025mmecomprehensiveevaluationbenchmark}), 
generation (GenEval~\citep{ghosh2023geneval},
DPG-Bench~\citep{hu2024ella})
and editing benchmarks (GEdit~\citep{liu2025step1x}). 
Unless otherwise stated, 
we conduct ablation studies on the ImageNet training set.

\textbf{Settings.}
UniWeTok adopts the architecture proposed in WeTok \citep{zhuang2025wetok} with a downsampling factor of 32 and a codebook size of $2^{128}$.
Images are randomly cropped to target sizes for training.
For ablation study, 
all models are trained for 250K steps with Adam \citep{Kingma2014AdamAM} and a consistent set of hyperparameters. 
For large-scale training, 
hyperparameters are individually tuned for each model to achieve optimal performance. 
For class-to-image generation and Unified MLLM,
we adopt the model architecture and training setting in BitDance~\citep{bitdance}.

\subsection{Ablation Studies}
\label{subsec:ablations}

\textbf{Training loss.}
We conduct ablation studies on the WeTok loss, the semantic distillation loss, and the prior loss. As shown in Tab.~\ref{tab:ablation_loss}, incorporating the semantic distillation loss significantly enhances the semantic extraction capability of the tokenizer. Furthermore, the introduction of the prior loss improves the tokenizer's performance on downstream generation while maintaining its reconstruction capability. 
Unexpectedly, 
the prior loss not only preserves the model's understanding capability but actually enhances it.
\begin{table}[!ht]
    \centering
    \caption{\textbf{Ablation study of semantic distillation and prior loss.} 
    We ablate the contribution of each loss. }
    \label{tab:ablation_loss}

    \resizebox{0.75\linewidth}{!}{
        \begin{tabular}{cc|c|ccccc|ccc}
            \toprule
            \multicolumn{2}{c|}{\textbf{Setting}} & \textbf{Gen.} & \multicolumn{5}{c|}{\textbf{Reconstruction}} & \multicolumn{3}{c}{\textbf{Zero-shot Acc (\%)}} \\
            PPD & GAP & gFID$\downarrow$ & rFID$\downarrow$ & PSNR$\uparrow$ & SSIM$\uparrow$ & LPIPS(A) & LPIPS(V) & Top-1 & Top-5 & Top-10 \\
            \midrule
            \xmark & \xmark & 2.38 & 1.33 & \textbf{22.67} & \textbf{0.66} & \textbf{0.11} & \textbf{0.20} & - & - & - \\
            \cmark & \xmark & 2.66 & \textbf{1.12} & 22.32 & 0.65 & 0.12 & 0.21 & 46.89 & 73.66 & 81.11 \\
            \rowcolor{bestblue} \cmark & \cmark & \textbf{2.35} & 1.18 & 22.27 & 0.63 & 0.12 & 0.22 & \textbf{48.77} & \textbf{75.62} & \textbf{83.84} \\
            \bottomrule
        \end{tabular}
    }
\end{table}

\textbf{SigLu activation.}
As shown in Tab.~\ref{tab:distill_strat}, 
the GQ poses a significant challenge for semantic extraction. 
Post distillation almost fails to converge. 
Pre distillation stabilizes convergence, 
while it fails to ensure semantics in the discrete latents. 
Our proposed SigLu activation constrains the feature space, 
enabling effective post distillation.
\begin{table}[!ht]
    \centering
    \begin{minipage}[b]{0.38\linewidth}
        \centering
        \tablestyle{2pt}{1.2}
        \caption{\textbf{SigLu activation ablation.} SigLu activation makes post-distillation effective.}
        \label{tab:distill_strat}

        \setlength{\tabcolsep}{4pt}
        
        \resizebox{\linewidth}{!}{
            \begin{tabular}{l|ccc}
                \toprule
                Method & Top-1 & Top-5 & Top-10 \\
                \midrule
                \rowcolor{rowgray}
                Pre Distill & 55.26 & 79.55 & 85.52 \\
                Post Distill & 0.10 & 0.53 & 0.95 \\
                \rowcolor{bestblue} \textbf{SigLu + Post} & \textbf{41.51} & \textbf{69.51} & \textbf{77.73} \\
                \bottomrule
            \end{tabular}
        }
    \end{minipage}
    \hspace{0.5cm}
    \begin{minipage}[b]{0.35\linewidth}
        \centering
        \tablestyle{4pt}{1.2}
        \caption{\textbf{Pre-Post Distillation ablation.} Pre-Post yields the best performance.}
        \label{tab:distill_loss}        
        \resizebox{\linewidth}{!}{
            \begin{tabular}{cc|ccc}
                \toprule
                \multicolumn{2}{c|}{\textbf{Distillation Loss}} & \multicolumn{3}{c}{\textbf{Zero-shot Acc (\%)}} \\
                \textsc{Pre} & \textsc{Post} & Top-1 & Top-5 & Top-10 \\
                \midrule
                \cmark & \xmark & 0.10 & 0.52 & 1.01 \\
                \xmark & \cmark & 41.51 & 69.51 & 77.73 \\
                \rowcolor{bestblue} \cmark & \cmark & \textbf{51.32} & \textbf{76.70} & \textbf{83.26} \\
                \bottomrule
            \end{tabular}
        }
    \end{minipage}
\end{table}

\textbf{Pre-Post Distillation.}
As shown in Tab.~\ref{tab:distill_loss},we demonstrate the effectiveness of our pre-post distillation strategy.
The combination of pre and post distillation yields the best performance, reaching 51.32\% zero-shot Top-1 accuracy.

\textbf{Bottleneck channel.}
As shown in Tab.~\ref{tab:channel}, 
we ablate channel widths of the bottlenecks.
Doubling the channel significantly lowers rFID (from 1.58 to 1.12) and improves semantic accuracy by over 7\%, 
proving that a wider bottleneck is essential for visual compression and semantic extraction.
\begin{table}[!ht]
    \centering
    \caption{\textbf{Bottleneck channel ablation.} Extending channel width of bottleneck significantly improves reconstruction fidelity and semantic extraction capability.}
    \label{tab:channel}
        \resizebox{0.8\linewidth}{!}{
            \begin{tabular}{l|ccccc|ccc}
                \toprule
                & \multicolumn{5}{c|}{\textbf{Reconstruction}} & \multicolumn{3}{c}{\textbf{Zero-shot Acc (\%)}} \\
                \multirow{-2}{*}{\textbf{Setting}} & rFID$\downarrow$ & PSNR$\uparrow$ & SSIM$\uparrow$ & LPIPS(A) & LPIPS(V) & Top-1 & Top-5 & Top-10 \\
                \midrule
                Single Channel & 1.58 & 21.89 & 0.63 & 0.13 & 0.23 & 39.45 & 68.00 & 76.77 \\
                \rowcolor{bestblue} \textbf{Double Channel} & \textbf{1.12} & \textbf{22.32} & \textbf{0.65} & \textbf{0.12} & \textbf{0.21} & \textbf{46.89} & \textbf{73.66} & \textbf{81.11} \\
                \bottomrule
            \end{tabular}
    }
\end{table}

\textbf{Generative-Aware Prior.}
Tab.~\ref{tab:ar_prior} shows that while reconstruction metrics remain stable,
generative quality (gFID) improves notably from 2.66 to 2.38 with GAP and Query token, 
confirming that regularizing the latent space benefits downstream generation task.
\begin{table}[!ht]
    \centering
            \caption{\textbf{Generative-Aware Prior ablation.} 
            }
        \label{tab:ar_prior}
        \resizebox{0.85\linewidth}{!}{
            \begin{tabular}{cc|c|ccccc|ccc}
                \toprule
                \multicolumn{2}{c|}{\textbf{Method}}
                & \textbf{Gen.} & \multicolumn{5}{c|}{\textbf{Reconstruction}} & \multicolumn{3}{c}{\textbf{Zero-shot Acc (\%)}} \\
                \textsc{GAP} & \textsc{Query} & gFID$\downarrow$ & rFID$\downarrow$ & PSNR$\uparrow$ & SSIM$\uparrow$ & LPIPS(A) & LPIPS(V) & Top-1 & Top-5 & Top-10 \\
                \midrule
                \xmark & \xmark & 2.66 & \textbf{1.12} & \textbf{22.32} & \textbf{0.65} & \textbf{0.12} & \textbf{0.21} & 46.89 & 73.66 & 81.11 \\
                \cmark & \xmark & 3.89 & 1.16 & 22.11 & 0.63 & 0.13 & 0.22 & 48.47 & 74.80 & 82.23 \\
                \rowcolor{bestblue} \cmark & \cmark & \textbf{2.38} & 1.18 & 22.27 & 0.63 & 0.12 & 0.22 & \textbf{48.77} & \textbf{75.62} & \textbf{83.84} \\
                \bottomrule
            \end{tabular}
    }
\end{table}

\textbf{Training configurations on DataComp-1B.}
Tab.~\ref{tab:head_batch} shows that attention head is better than linear head.
Moreover, 
scaling the batch size tripling the zero-shot accuracy.
\begin{table}[!ht]
    \centering
    \tablestyle{4.5pt}{1.1}
    \caption{\textbf{Training configuration ablations on DataComp-1B.} 
    Attention-based semantic head and large batch size are critical for model convergence on general-domain dataset.}
    \label{tab:head_batch}
    \resizebox{0.8\linewidth}{!}{
    \begin{tabular}{l|ccccc|ccc}
    \toprule
    \multirow{2}{*}{\textbf{Setting}} & \multicolumn{5}{c|}{\textbf{Reconstruction}} & \multicolumn{3}{c}{\textbf{Zero-shot Acc (\%)}} \\
    ~ & rFID $\downarrow$ & PSNR $\uparrow$ & SSIM $\uparrow$ & LPIPS(A) & LPIPS(V) & Top-1 & Top-5 & Top-10 \\
    \midrule
    \rowcolor{rowgray}
    \multicolumn{9}{l}{\textit{Semantic Head Architecture}} \\
    Linear Head & \textbf{2.91} & 21.96 & 0.63 & 0.13 & 0.24 & 3.89 & 13.85 & 21.93 \\
    \rowcolor{bestblue}
    \textbf{Attention Head} & 3.10 & \textbf{22.02} & \textbf{0.63} & \textbf{0.13} & \textbf{0.24} & \textbf{4.09} & \textbf{14.39} & \textbf{23.31} \\
    \midrule
    \rowcolor{rowgray}
    \multicolumn{9}{l}{\textit{Training Batch Size}} \\
    Batch Size 128 & 3.10 & 22.02 & 0.63 & 0.13 & 0.24 & 4.09 & 14.39 & 23.31 \\
    \rowcolor{bestblue}
    \textbf{Batch Size 1024} & \textbf{1.75} & \textbf{22.56} & \textbf{0.65} & \textbf{0.12} & \textbf{0.22} & \textbf{11.69} & \textbf{31.38} & \textbf{43.34} \\
    \bottomrule
    \end{tabular}}
\end{table}

\textbf{Hybrid architecture.}
On Datacomp-1B,
Tab.~\ref{tab:hybrid} validates our hybrid backbone.
CNNs excel at texture (rFID 1.75) but lack semantics, while Transformers capture semantics (Top-1 26.09\%) but struggle with detail. Our hybrid design achieves the best of both worlds (rFID 1.35, Top-1 35.41\%).
\begin{table}[!ht]
    \centering
    \tablestyle{6pt}{1.1}
    \caption{\textbf{Architecture ablation.} The hybrid backbone successfully synergizes the local inductive bis of convolution with the global understanding capability of attention.}
    \label{tab:hybrid}
    \resizebox{0.8\linewidth}{!}{
    \begin{tabular}{l|c c c c c|c c c}
        \toprule
         & \multicolumn{5}{c|}{\textbf{Reconstruction}} & \multicolumn{3}{c}{\textbf{Zero-shot Acc (\%)}} \\
        \multirow{-2}{*}{\textbf{Backbone}} & rFID $\downarrow$ & PSNR $\uparrow$ & SSIM $\uparrow$ & LPIPS(A) & LPIPS(V) & Top-1 & Top-5 & Top-10 \\
        \midrule
        CNN Only & 1.75 & \textbf{22.56} & 0.65 & \textbf{0.12} & \textbf{0.22} & 11.69 & 31.38 & 43.34 \\
        Transformer Only & 3.38 & 22.02 & 0.65 & 0.14 & 0.26 & 26.09 & 52.77 & 64.28 \\
        \rowcolor{bestblue}
        \textbf{Hybrid (Ours)} & \textbf{1.35} & 22.54 & \textbf{0.66} & 0.13 & 0.22 & \textbf{35.41} & \textbf{63.91} & \textbf{73.50} \\
        \bottomrule
    \end{tabular}}
\end{table}

\textbf{Three-stage training pipeline.}
As shown in Fig.~\ref{fig:train_pipe_ablation},
compared to stage 1, the model trained in stage 2 supports variable-resolution image reconstruction. Furthermore, the model trained in stage 3 demonstrates significantly improved performance in processing faces and text.

Implementation details could be found in App.~\ref{app:imple_detail}.

\subsection{Comparison with State-of-the-Art}
\label{subsec:comp_sota}

\textbf{Visual Generation on ImageNet.}
We first evaluate the reconstruction performance of UniWeTok on the ImageNet 50K validation dataset. 
As shown in Tab.~\ref{tab:comp_imagenet_recon}, 
UniWeTok compresses a $256 \times 256$ resolution image into only \textbf{64 tokens}, 
which represents a \textbf{75\%} reduction in token count compared to other mainstream tokenizers.
Following the BitDance setting, 
we employ UniWeTok as the visual tokenizer for autoregressive modeling.
As shown in Tab.~\ref{tab:comp_imagenet_gen}, UniWeTok-H achieves a state-of-the-art FID of \textbf{1.38}. 
Remarkably, 
our model requires training on only \textbf{33B} tokens and generates just 64 tokens during inference, significantly outperforming various models that demand extensive training scales and incur higher computational costs during inference.
\begin{table}[t]
    \centering
    \renewcommand\arraystretch{1.1}
    \caption{\textbf{Reconstruction evaluation on $\mathbf{256\times256}$ ImageNet 50K validation set.}
    All models are trained on ImageNet.
    UniWeTok achieves SOTA results with $32\times$ downsampling ratio.}
    \resizebox{0.75\linewidth}{!}{
        \begin{tabular}{lcccccc}
        \toprule
        \multirow{2}{*}{\textbf{Method}} & \multirow{2}{*}{\textbf{Tokens}$\downarrow$} & \multirow{2}{*}{\textbf{Ratio}$\uparrow$} & \textbf{Codebook} & \multirow{2}{*}{\textbf{rFID}$\downarrow$} & \multirow{2}{*}{\textbf{PSNR}$\uparrow$} & \textbf{Codebook} \\
         & & & \textbf{Size} & & & \textbf{Usage}$\uparrow$ \\
        \midrule
        VQGAN~\citep{Esser2020TamingTF} & $16\times16$ & 16 & $16384$ & 4.99 & 20.00 & $-$ \\
        SD-VQGAN~\citep{rombach2022sd} & $16\times16$ & 16 & $16384$ & 5.15 & $-$ & $-$ \\
        MaskGIT~\citep{Chang2022MaskGITMG} & $16\times16$ & 16 & $1024$ & 2.28 & $-$ & $-$ \\
        ReVQ~\citep{zhang2025quantize} & $16\times16$ & 16 & $65536$ & 2.57 & 21.69 & $-$ \\
        LlamaGen~\citep{Sun2024AutoregressiveMB} & $16\times16$ & 16 & $16384$ & 2.19 & 20.79 & 97\% \\
        ReVQ~\citep{zhang2025quantize} & $16\times16$ & 16 & $2^{18}$ & 2.05 & 21.96 & $-$ \\
        
        TiTok~\citep{Yu2024AnII} & $256$ & 16 & $4096$ & 1.66 & 20.01 & 100\% \\
        FlexTok~\citep{bachmann2025flextok} & $256$ & 16 & $64000$ & 1.45 & 18.53 & $-$ \\
        VAR~\citep{Tian2024VisualAM} & $16\times16$ & 16 & $4096$ & $-$ & 21.30 & 97\% \\
        IBQ~\citep{shi2025scalableimagetokenizationindex} & $16\times16$ & 16 & $16384$ & 1.37 & 22.35 & 96\% \\
        
        Open-MAGVIT2~\citep{Luo2024OpenMAGVIT2AO} & $16\times16$ & 16 & $2^{18}$ & 1.17 & 22.64 & 100\% \\ 
        IBQ~\citep{shi2025scalableimagetokenizationindex} & $16\times16$ & 16 & $2^{18}$ & 1.00 & 20.30 & 84\% \\

        FlowMo-Lo~\citep{sargent2025flow} & $256$ & 16 & $2^{18}$ & 0.95 & 22.07 & $-$ \\
        VFMTok~\citep{zheng2025vision} & $256$ & 16 & $16384$ & 0.89 & $-$ & 100\% \\
        
        GigaTok~\citep{xiong2025gigatokscalingvisualtokenizers} & $256$ & 16 & $16384$ & 0.79 & 21.65 & $-$ \\
        
        AliTok~\citep{wu2025alitok} & $273$ & 15.5 & $4096$ & 0.84 & $-$ & $-$ \\ 

        \rowcolor{bestblue}
        \textbf{UniWeTok (Ours)} & $\mathbf{8\times8}$ & \textbf{32} & $\mathbf{2^{128}}$ & \textbf{0.79} & \textbf{23.26} & \textbf{100\%} \\ 
    \bottomrule
    \end{tabular}}
    \label{tab:comp_imagenet_recon}
\end{table}
\begin{table}[!ht]
\caption{\textbf{Comparison of class-conditional image generation on ImageNet 256$\times$256.} UniWeTok achieves superior performance while employing standard causal autoregressive modeling.}
\label{tab:comp_imagenet_gen}
\centering
\resizebox{0.75\linewidth}{!}{
\begin{tabular}{lcc|c|cc|cc}
\toprule
\textbf{Method} & \textbf{\makecell{Train\\Tokens$\downarrow$}} & \textbf{\makecell{Infer\\Tokens$\downarrow$}} & \textbf{\#Params} & \textbf{FID$\downarrow$} & \textbf{IS$\uparrow$} & \textbf{Pre.$\uparrow$} & \textbf{Rec.$\uparrow$} \\
\midrule
\multicolumn{8}{l}{\textit{Continuous Tokens}} \\
\addlinespace[2.5pt]
~ DiT-XL/2~\cite{dit}         & 459B & 256 & 675M  & 2.27 & 278.2 & 0.83 & 0.57 \\
~ SiT-XL/2~\cite{ma2024sit}          & 459B & 256 & 675M  & 2.06 & 277.5 & 0.83 & 0.59 \\
~ MDTv2~\cite{gao2023mdtv2}         & 131B & 256 & 675M  & 1.58 & 314.7 & 0.79 & \textbf{0.65} \\
~ REPA~\cite{repa}  & 262B & 256 & 675M  & 1.42 & 305.7 & 0.80 & \textbf{0.65} \\
~ MAR-B~\cite{li2024autoregressive}           & 262B & 256 & 208M  & 2.31 & 281.7 & 0.82 & 0.57 \\
~ MAR-L~\cite{li2024autoregressive}             & 262B & 256 & 479M  & 1.78 & 296.0 & 0.81 & 0.60 \\
~ MAR-H~\cite{li2024autoregressive}             & 262B & 256 & {943M} & {1.55} & \textbf{303.7} & {0.81} & {0.62} \\
~ NiT-XL~\cite{wang2025native}              & 197B & \textbf{64} & {675M} & {2.03} & 265.26 & {0.80} & {0.62} \\

\midrule
\multicolumn{8}{l}{\textit{Discrete Tokens}} \\
\addlinespace[2pt]
~ LlamaGen-L~\cite{sun2024autoregressive}       & 98B & 256 & 343M   & 3.07 & 256.1 & 0.83 & 0.52 \\
~ LlamaGen-XL~\cite{sun2024autoregressive}       & 98B & 256 & 775M   & 2.62 & 244.1 & 0.80 & 0.57 \\
~ LlamaGen-XXL~\cite{sun2024autoregressive}     & 98B & 256 & 1.4B   & 2.34 & 253.9 & 0.80 & 0.59 \\
~ RandAR-L~\cite{pang2025randar}       & 98B & 256 & 343M   & 2.55 & 288.8 & 0.81 & 0.58 \\
~ RandAR-XL~\cite{pang2025randar}        & 98B & 256 & 775M   & 2.22 & 314.2 & 0.80 & 0.60 \\
~ RandAR-XXL~\cite{pang2025randar}        & 98B & 256 & 1.4B   & 2.15 & 322.0 & 0.79 & 0.62 \\
~ RAR-L~\cite{yu2025randomized}      & 131B & 256 & 461M   & 1.70 & 299.5 & 0.81 & 0.60 \\
~ RAR-XL~\cite{yu2025randomized}      & 131B & 256 & 955M   & 1.50 & 306.9 & 0.80 & 0.62 \\
~ RAR-XXL~\cite{yu2025randomized}      & 131B & 256 & 1.5B   & 1.48 & \textbf{326.0} & 0.80 & 0.63 \\
~ OpenMAGVIT2-XL~\cite{luo2024open}        & 115B & 256 & 1.5B   & 2.33 & 271.8 & \textbf{0.84} & 0.54 \\
~ MAGVIT-v2~\cite{yulanguage}        & 88B & 256 & 307M & 1.78 & 319.4 & - & - \\
~ VAR-d20~\cite{tian2024visual}        & 82B & 256 & 600M  & 2.57 & 302.6 & 0.83 & 0.56 \\
~ VAR-d30~\cite{tian2024visual}        & 115B & 256 & 2B   & 1.92 & 323.1 & 0.82 & 0.59 \\
~ WeTok-AR-XL~\cite{zhuang2025wetok}        & 328B & 256 & 1.5B & 2.31 & 276.6 & \textbf{0.84} & 0.55 \\
\rowcolor{bestblue}
~ \textbf{UniWeTok-B}        & \textbf{33B} & \textbf{64} & 242M   & 2.35 & 284.47 & 0.80 & 0.59 \\
\rowcolor{bestblue}
~ \textbf{UniWeTok-L}        & \textbf{33B} & \textbf{64} & 527M   & 1.68 & 288.56 & 0.80 & 0.62 \\
\rowcolor{bestblue}
~ \textbf{UniWeTok-H}       & \textbf{33B} & \textbf{64} & 1.0B   & \textbf{1.38} & 284.34 & 0.80 & 0.63 \\
\bottomrule
\end{tabular}}
\end{table}

\textbf{Unified MLLM.}
We first compared the reconstruction performance of tokenizers trained on general-domain dataset.
As shown in Tab.~\ref{tab:comp_genneral_recon},
our UniWeTok outperforms most state-of-the-art general tokenizers in reconstruction performance while utilizing only \textbf{25\%} of the visual token count. 
This reduction in sequence length allows for larger batch sizes in the subsequent training of unified MLLMs, 
significantly enhancing training efficiency.
Given UniWeTok's capability to support image compression and reconstruction at variable resolutions,
we employ \textbf{native resolutions} for Unified MLLM training. 
This ensures that the inherent spatial information of the data is preserved without the distortions introduced by resizing or cropping.
Following Emu3, 
after unified pretraining,
we finetune the Unified MLLM into UniWeTok-Gen,
UniWeTok-Edit and UniWeTok-Chat for better downstream performance.
As shown in Tab.~\ref{tab:comp_general_gen}, 
UniWeTok-Gen demonstrates superior performance on image generation benchmarks,
outperforming various diffusion-based models.
As shown in Tab.~\ref{tab:comp_general_und},
UniWeTok-Chat presents competitive understanding performance across a broad range of benchmarks. 
In Tab.~\ref{tab:comp_general_edit},
UniWeTok-Edit surpasses the diffusion model on the image editing task as an autoregressive model at a similar parameter scale for the first time.
The visualization results can be seen in Fig.~\ref{fig:teaser},\ref{fig:vis_edit1},\ref{fig:vis_edit2},\ref{fig:vis_edit3}.
\begin{table}[t]
    \centering
    \renewcommand\arraystretch{1.1}
    \caption{\textbf{Zero-shot reconstruction comparison on ImageNet and MS-COCO val2017 validation set.} 
    Our UniWeTok achieves the best performance on different resolution settings.
    }
    \resizebox{0.75\linewidth}{!}{
        \begin{tabular}{lcccccccc}
        \toprule
        \multirow{2}{*}{\textbf{Method}} & \multirow{2}{*}{\textbf{Ratio}$\downarrow$} & \multicolumn{3}{c}{\textbf{MS-COCO 2017}} & & \multicolumn{3}{c}{\textbf{Imagenet-1k}} \\
        \cmidrule{3-5} \cmidrule{7-9}
         & & rFID$\downarrow$ & PSNR$\uparrow$ & SSIM$\uparrow$ & & rFID$\downarrow$ & PSNR$\uparrow$ & SSIM$\uparrow$ \\
        \midrule
        \multicolumn{9}{c}{\textit{Resize $256 \times 256$}} \\
        \midrule
        Cosmos~\citep{Agarwal2025CosmosWF} & 16 & 11.97 & 19.22 & 0.48 & & 4.57 & 19.93 & 0.49 \\ 
        Show-o~\citep{Xie2024ShowoOS} & 16 & 9.26 & 20.90 & 0.59 & & 3.50 & 21.34 & 0.59 \\
        WeTok~\citep{zhuang2025wetok} & \textbf{32} & 8.94 & 20.31 & 0.55 & & 3.49 & 20.77 & 0.55 \\
        \small{Open-MAGVIT2-I-PT~\citep{Luo2024OpenMAGVIT2AO}} & 16 & 7.93 & 22.21 & 0.62 & & 2.55 & 22.21 & 0.62 \\
        LlamaGen~\citep{Sun2024AutoregressiveMB} & 16 & 8.40 & 20.28 & 0.55 & & 2.47 & 20.65 & 0.54 \\

        WeTok~\citep{zhuang2025wetok} & 16 & 6.55 & 21.99 & 0.63 &  & 1.58 & 22.38 & 0.62 \\
        BSQ~\citep{zhao2024image} & 16 & - & - & - & & 3.81 & 24.12 & 0.66 \\
        QLIP-B~\citep{Zhao2025QLIPTV} & 16 & - & - & - & & 3.21 & 23.16 & 0.63 \\
        QLIP-L~\citep{Zhao2025QLIPTV} & 14 & - & - & - & & 1.46 & \textbf{25.36} & \textbf{0.69} \\
        TokenFlow~\citep{Zhao2025QLIPTV} & 16 & - & - & - & & 1.37 & 21.41 & \textbf{0.69} \\
        \rowcolor{bestblue}
        \textbf{UniWeTok (Ours)} & \textbf{32} & \textbf{6.18} & \textbf{22.58} & \textbf{0.66} &  & \textbf{1.18} & 22.97 & 0.66 \\
        \midrule
        
        \multicolumn{9}{c}{\textit{Original Resolution}} \\
        \midrule
        WeTok~\citep{zhuang2025wetok} & \textbf{32} & 8.94 & 20.31 & 0.55 & & 3.49 & 20.77 & 0.55 \\
        Cosmos~\citep{Agarwal2025CosmosWF} & 16 & 7.23 & 20.45 & 0.53 & & 2.52 & 20.49 & 0.52 \\
        \small{Open-MAGVIT2-I-PT~\citep{Luo2024OpenMAGVIT2AO}} & 16 & 6.65 & 21.61 & 0.57 & & 1.39 & 21.74 & 0.56 \\
        SD-VAE 1.x~\citep{rombach2022high} & 8 & 5.94 & 21.68 & 0.64 & & 1.35 &  21.99 & 0.63 \\
        WeTok~\citep{zhuang2025wetok} & 16 & \textbf{5.30} & 21.94 & 0.59 & & \textbf{0.81} & 21.99 & 0.58 \\
        \rowcolor{bestblue}
        \textbf{UniWeTok (Ours)} & \textbf{32} & 6.46 & \textbf{22.29} & \textbf{0.65} & & 1.25 & \textbf{22.66} & \textbf{0.65} \\

    \bottomrule
    \end{tabular}}
    \label{tab:comp_genneral_recon}
\end{table}
\begin{table*}[!ht]
\centering
\caption{
\textbf{Comparisons of visual generation quality on GenEval and DPG-Bench.}
}
\setlength{\tabcolsep}{1pt}
\resizebox{1\linewidth}{!}{
    \begin{tabular}{lr|ccccccc|cccccc}
    \toprule
    \multirow{2}{*}{\textbf{Method}} & \multirow{2}{*}{\textbf{\# Params}} & \multicolumn{7}{c}{\textbf{GenEval}~\citep{ghosh2023geneval}} & \multicolumn{6}{c}{\textbf{DPG-Bench}~\citep{hu2024ella}} \\
    \cmidrule(lr){3-9} \cmidrule(lr){10-15} 
    & & \textbf{Single Obj.} & \textbf{Two Obj.} & \textbf{Counting} & \textbf{Colors} & \textbf{Position} & \textbf{Color Attri.} & \textbf{Overall$\uparrow$} & \textbf{Global} & \textbf{Entity} & \textbf{Attribute} & \textbf{Relation} & \textbf{Other} & \textbf{Overall$\uparrow$} \\
    \midrule
    \multicolumn{15}{c}{\textit{Diffusion-based Model}} \\
    \midrule
    SDv1.5~\citep{rombach2022high} & 0.9B & 0.97 & 0.38 & 0.35 & 0.76 & 0.04 & 0.06 & 0.43 & 74.63 & 74.23 & 75.39 & 73.49 & 67.81 & 63.18 \\
    PixArt-$\alpha$~\citep{chen2023pixart-alpha} & 0.6B & 0.98 & 0.50 & 0.44 & 0.80 & 0.08 & 0.07 & 0.48 & 74.97 & 79.32 & 78.60 & 82.57 & 76.96 & 71.11 \\
    SDv2.1~\citep{rombach2022high} & 0.9B & 0.98 & 0.51 & 0.44 & 0.85 & 0.07 & 0.17 & 0.50 & - & - & - & - & - & - \\
    SDXL~\citep{podell2023sdxl} & 2.6B & 0.98 & 0.74 & 0.39 & 0.85 & 0.15 & 0.23 & 0.55 & 83.27 & 82.43 & 80.91 & 86.76 & 80.41 & 74.65 \\
    Playground v2.5~\citep{li2024playground} & 2.6B & - & - & - & - & - & - & - & 83.06 & 82.59 & 81.20 & 84.08 &83.50 & 75.47 \\
    Hunyuan DiT~\citep{li2024hunyuan} & 1.5B & - & - & - & - & - & - & - & 84.59 & 80.59 & 88.01 & 74.36 & 86.41 & 78.87 \\
    PixArt-$\Sigma$~\citep{chen2024pixart} & 0.6B & - & - & - & - & - & - & - & 86.89 & 82.89 & 88.94 & 86.59 & 87.68 & 80.54 \\
    DALLE3~\citep{betker2023dalle3} & - & 0.96 & 0.87 & 0.47 & 0.83 & 0.43 & 0.45 & 0.67 & 90.97 & 89.61 & 88.39 & 90.58 & 89.83 & 83.50 \\
    SD3-Medium~\citep{esser2024sd3} & 2B & 0.99 & 0.94 & 0.72 & 0.89 & 0.33 & 0.60 & 0.74 & 87.90 & 91.01 & 88.83 & 80.70 & 88.68 & 84.08 \\
    SANA-1.5~\citep{xie2025sana} & 4.8B & 0.99 & 0.93 & 0.86 & 0.84 & 0.59 & 0.65 & \textbf{0.81} & - & - & - & - & - & 84.70 \\
    \midrule
    \multicolumn{15}{c}{\textit{Autoregressive-based Model}} \\
    \midrule
    Chameleon~\citep{team2024chameleon} & 7B & - & - & - & - & - & - & 0.39 & - & - & - & - & - & - \\
    LlamaGen~\citep{li2024autoregressive} & 0.8B & 0.71 & 0.34 & 0.21 & 0.58 & 0.07 & 0.04 & 0.32 & 81.76 & 75.43 & 76.17 & 84.76 & 58.40 & 64.84 \\
    EMU3-Gen~\citep{wang2024emu3} & 8B & 0.98 & 0.71 & 0.34 & 0.81 & 0.17 & 0.21 & 0.54 & 85.21 & 86.68 & 86.84 & 90.22 & 83.15 & 80.60 \\
    TokenFlow~\citep{Qu2024TokenFlowUI} & 13B & 0.97 & 0.66 & 0.40 & 0.84 & 0.17 & 0.26 & 0.55 & 78.72 & 79.22 & 81.29 & 85.22 & 71.20 & 73.38 \\
    Janus~\citep{wu2025janus} & 1.3B & 0.97 & 0.68 & 0.30 & 0.84 & 0.46 & 0.42 & 0.61 & 82.33 & 87.38 & 87.70 & 85.46 & 86.41 & 79.68 \\
    SimpleAR~\citep{wang2025simplear} & 1.5B & - & 0.90 & - & - & 0.28 & 0.45 & 0.63 & 87.97 & - & - & 86.33 & - & 81.97 \\
    Transfusion~\citep{zhou2024transfusion} & 7B & - & - & - & - & - & - & 0.63 & - & - & - & - & - & - \\
    NextStep-1~\citep{team2025nextstep} & 14B & - & - & - & - & - & - & 0.73 & - & - & - & - & - & 85.28 \\
    Harmon~\citep{wu2025harmonizing} & 1.5B & 0.99 & 0.86 & 0.66 & 0.85 & 0.74 & 0.48 & 0.76 & - & - & - & - & - & - \\
    Infinity~\citep{han2025infinity} & 8B & & & & & & & 0.79 & & & & & & 86.6
    \\
    Janus-Pro~\citep{chen2025janus} & 7B & 0.99 & 0.89 & 0.59 & 0.90 & 0.79 & 0.66 & 0.80 & 86.90 & 88.90 & 89.40 & 89.32 & 89.48 & 84.19 \\
    \rowcolor{bestblue}
    \textbf{UniWeTok-Gen (Ours)} & 8B & 0.99 & 0.94 & 0.43 & 0.90 & 0.78 & 0.81 & \textbf{0.81} & 87.54 & 91.99 & 90.72 & 92.92 & 91.11 & \textbf{86.63} \\
    \bottomrule
    \end{tabular}
}
\label{tab:comp_general_gen}
\end{table*}
\begin{table*}[!ht]
    \centering
    \setlength{\tabcolsep}{0.075cm}
    \caption{\textbf{Comparison on multimodal understanding benchmarks.}
    }
    \resizebox{\linewidth}{!}{
        \begin{tabular}{llcccccccccccc}
            \toprule
            Method & Pretrained-LLM
            & SEEDB & POPE
            & VQAv2 & GQA & SQA & TQA 
            & CQA & AI2D 
            & RWQA & MMMU & MME-P & MME-S
            \\
            \midrule
            \multicolumn{11}{l}{\textbf{\textit{Understanding Only Model}}}\\
            \midrule
            InstructBLIP~\citep{dai2023instructblip}
            & Vicuna-7B
            & 53.4 & -- 
            & --   & 49.2 & 60.5 & 50.1 
            & 12.5 & 33.8 & 37.4 & 30.6 & 1212.8 & --
            \\
            IDEFICS-9B~\citep{laurenccon2023obelics}
            & LLaMA-7B
            & -- & -- 
            & 50.9 & 38.4 & --   & 25.9 
            & -- & 42.2 & 42.1 & 18.4 & -- & --
            \\
            LLaVA-1.5~\citep{liu2024improved}
            & Vicuna-7B
            & 64.3 & 85.9
            & 78.5 & 62.0 & 66.8 & 46.1 & 18.2 & 54.8 & 54.8 & 35.3 & 1510.7 & --
            \\
            InternVL-Chat~\citep{chen2024far} 
            & Vicuna-7B
            & -- & 86.4
            & 79.3 & 62.9 & --  & 57.0 
            & -- & --   & --  & -- & 1298.5 & --
            \\
            mPLUG-Owl2~\citep{ye2024mplug}
            &LLaMA2-7B
            & 57.8 & 86.2
            & 79.4 & 56.1 & 68.7 & 58.2 
            & 22.8 & 55.7 & 50.3 & 32.7 & -- & --
            \\
            LLaVA-1.6(HD)~\citep{llava1.6}
            &Vicuna-7B
            & 64.7 & 86.5	
            & 81.8 & 64.2 & 70.2 & 64.9 
            & 54.8 & 66.6 & 57.8 & 35.1 & -- & 1778.0
            \\
            VILA~\citep{lin2024vila}
            &LLaMA2-7B
            & 61.1 & 85.5
            & 80.8 & 63.3 & 73.7 & 66.6 
            & -- & --   & --   & -- & 1533.0 & -- 
            \\
            \midrule
            \multicolumn{11}{l}{\textbf{\textit{Unified Model}}}
            \\
            \midrule
            Fuyu-8B(HD)~\citep{fuyu-8b}
            & Persimmon-8B
            & -- & 74.1
            & 74.2 & --   & --   & -- 
            & -- & 64.5 & --   & 27.9  & -- & --
            \\
            Chameleon-MT-34B~\citep{team2024chameleon}
            &-- & --   & --
            & 69.6 & --   & --   & -- & -- & --   & --   & -- & --  & --
            \\
            LWM-7B~\citep{liu2024world} 
            &--
            & -- & 75.2
            & 55.8 & 44.8 & 47.7 & -- & -- 
            & --   & --   & -- & -- & --  
            \\
            Show-o~\citep{Xie2024ShowoOS} 
            &Phi-1.5-1.3B
            & -- & 73.8
            & 59.3 & 48.7 & -- & -- & -- 
            & --   & --   & 25.1 & 1097.2 & --  
            \\
            VILA-U~\citep{wu2024vila} 
            & LLaMA2-7B
            & 56.3 & 83.9
            & 75.3 & 58.3 & -- & 48.3 & -- 
            & --   & --   & -- & 1336.2 & --  
            \\
            Harmon~\citep{wu2024vila} 
            & Qwen2.5-1.5B
            & 67.1 & 87.6
            & -- & 58.9 & -- & -- & -- 
            & --   & --   & 38.9 & 1155.0 & 1476.0  
            \\
            EVE-7B(HD)~\citep{diao2024unveiling} 
            &Vicuna-7B
            & 56.8 & 85.0 & 78.6 
            & 62.6 & 64.9 & 56.8 
            & -- & --   & --   & -- & 1305.7 & --
            \\
            Emu3-Chat~\citep{wang2024emu3}
            & -- 
            & 68.2 & 85.2
            & 75.1 & 60.3 & 89.2 & 64.7 
            & 68.6 & 70.0 & 57.4 & 31.6 & -- & -- \\
            TokenFlow-L~\citep{Ma2025UniTokAU} & Vicuna-13B & 62.6 & 85.0 & 73.9 & 60.3 & -- & 54.1 & -- & 56.6 & 49.2 & 34.4 & 1365.4 & 1622.9 \\
            UniTok~\citep{Ma2025UniTokAU} & LLaMA-2-7B & -- & 83.2 & 61.1 & -- & -- & 51.6 & -- & -- & -- & -- & 1448.0 & -- \\
            \rowcolor{bestblue}
            UniWeTok-Chat
            & Qwen3-8B 
            & 69.3 & 85.6
            & 75.8 & 63.1 & 80.3 
            & 53.7 & 65.1 & 73.9 & 54.8 & 40.0 & 1415.7 & 1796.6
            \\
            \bottomrule
            \end{tabular}
        }
        \label{tab:comp_general_und}
\end{table*}

\begin{table}[!ht]
    \centering
    \renewcommand{\arraystretch}{1.1}
    
    \caption{\textbf{Comparison on GEdit-Bench.} G\_SC, G\_PQ, and G\_O refer to the metrics evaluated by GPT-4.1.}
    \label{tab:comp_general_edit}
    
    \resizebox{0.7\linewidth}{!}{
        \begin{tabular}{l|ccc|ccc}
        \toprule
        \multirow{2}{*}{\textbf{Model}} &
        \multicolumn{3}{c|}{\textbf{GEdit-EN (Full set)}} & 
        \multicolumn{3}{c}{\textbf{GEdit-CN (Full set)}} \\
        \cmidrule(lr){2-4} \cmidrule(lr){5-7}
        & G\_SC$\uparrow$ & G\_PQ$\uparrow$ & G\_O$\uparrow$ 
        & G\_SC$\uparrow$ & G\_PQ$\uparrow$ & G\_O$\uparrow$ \\
        \midrule
        
        \rowcolor{rowgray}
        \multicolumn{7}{l}{\textit{Private}} \\
        Gemini 2.0~\citep{gemini220250312} & 6.73 & 6.61 & 6.32 & 5.43 & 6.78 & 5.36 \\
        GPT-4o~\citep{openai2025chatgpt4o} & 7.85 & 7.62 & 7.53 & 7.67 & 7.56 & 7.30 \\
        \midrule
        
        \rowcolor{rowgray}
        \multicolumn{7}{l}{\textit{Open-Source Diffusion}} \\
        Instruct-Pix2Pix~\citep{Brooks2022InstructPix2PixLT} & 3.58 & 5.49 & 3.68 & - & - & - \\
        MagicBrush~\citep{zhang2023magicbrush} & 4.68 & 5.66 & 4.52 & - & - & - \\
        AnyEdit~\citep{yu2024anyedit} & 3.18 & 5.82 & 3.21 & - & - & - \\
        OmniGen~\citep{xiao2024omnigen} & 5.96 & 5.89 & 5.06 & - & - & - \\
        Step1X-Edit~\citep{liu2025step1xeditpracticalframeworkgeneral} & 7.09 & 6.76 & 6.70 & 7.20 & 6.87 & 6.86 \\
        BAGEL~\citep{deng2025emerging} & 7.36 & 6.83 & 6.52 & 7.34 & 6.85 & 6.50 \\
        \midrule
        \rowcolor{rowgray}
        \multicolumn{7}{l}{\textit{Open-Source Autoregressive}} \\
        \rowcolor{bestblue}
        \textbf{UniWeTok-Edit} & 5.86 & 5.89 & 5.09 & 5.78 & 5.92 & 5.11 \\
        \bottomrule
        \end{tabular}
    }
\end{table}

\section{Conclusion}
\label{sec:conclusion}
We presented UniWeTok, 
a unified discrete tokenizer designed to resolve the conflict between high-fidelity reconstruction,
multimodal understanding and generation. 
By integrating a convolution-attention hybrid backbone with the SigLu activation function,
UniWeTok successfully incorporates Pre-Post Distillation and Generative-Aware Prior, 
allowing a unified discrete tokenizer with massive codebook size $2^{128}$ to extract semantic concepts,
fine-grained texture details, and generative prior effectively.
Our UniWeTok establishes a robust and efficient baseline for future Unified MLLM works, 
suggesting that a single, 
well-optimized tokenizer is sufficient to address the complex challenges inherent in Unified MLLMs.

\clearpage

\bibliographystyle{plainnat}
\bibliography{main}

\clearpage

\beginappendix
\section{Model Architecture}
\label{app:model_arch}

\begin{figure*}[!ht]
    \centering
    \includegraphics[width=\linewidth]{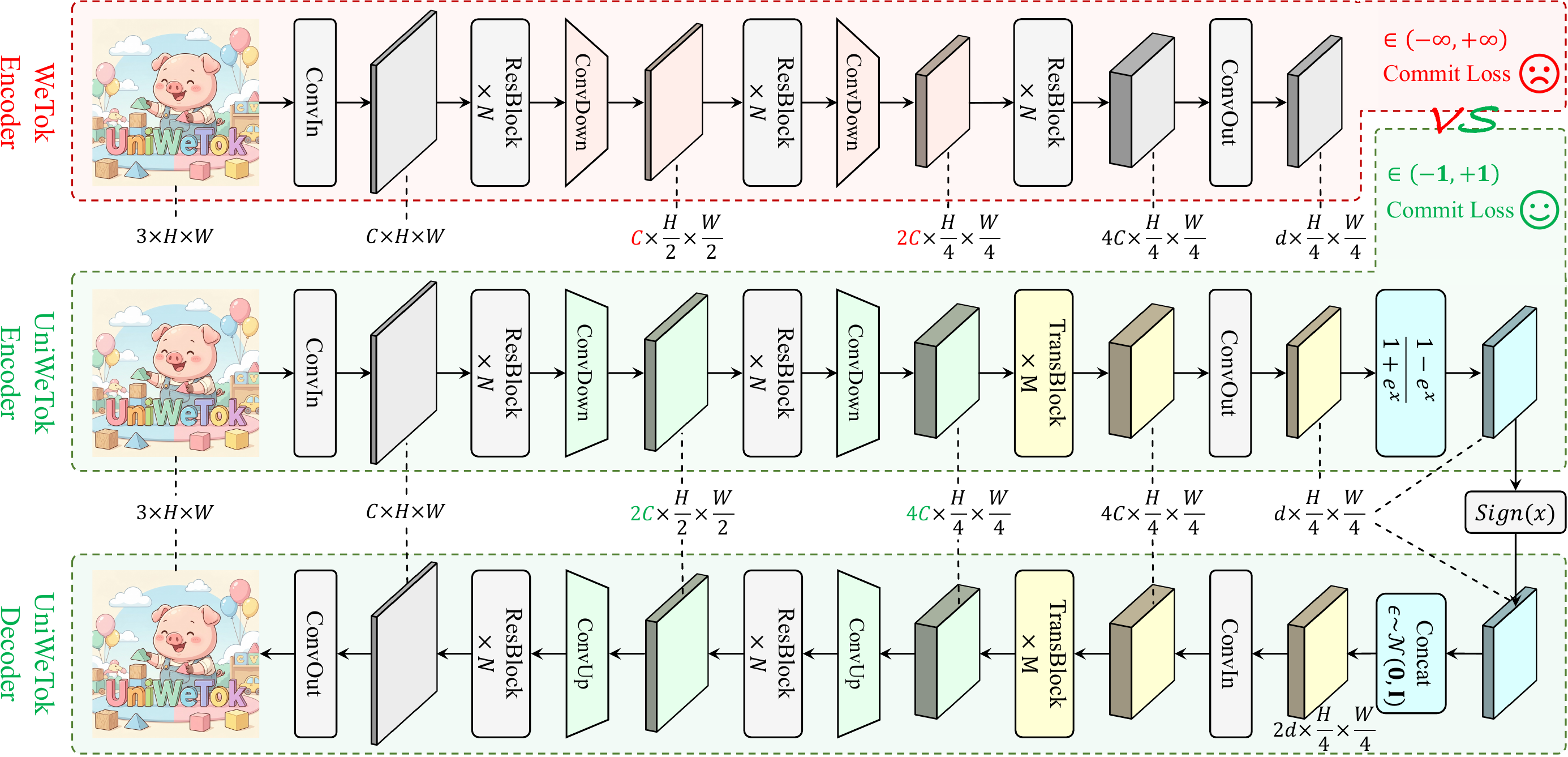}
    \caption{\textbf{Detail Illustration of the UniWeTok model architecture.}
    }
    \label{fig:model_arch}
\end{figure*}

\section{More Ablation Implementation Details}
\label{app:imple_detail}

\textbf{Training loss.}
As shown in Tab.~\ref{tab:imple_abla_wetok_loss_config},
\ref{tab:imple_abla_dis_loss_config},
\ref{tab:imple_abla_uniwetok_loss_config}.

\textbf{SigLu activation.}
As shown in Tab.~\ref{tab:imple_abla_nosiglu_pre},
\ref{tab:imple_abla_nosiglu_post},
\ref{tab:imple_abla_siglu_post}.

\textbf{Pre-Post Distillation.}
As shown in Tab.~\ref{tab:imple_abla_pre_dis},
\ref{tab:imple_abla_post_dis},
\ref{tab:imple_abla_pre_post_dis}.

\textbf{Bottleneck channel.}
As shown in Tab.~\ref{tab:imple_abla_single_ch},
\ref{tab:imple_abla_double_ch}.

\textbf{Generative-Aware Prior.}
As shown in Tab.~\ref{tab:imple_abla_no_prior_no_query},
\ref{tab:imple_abla_prior_no_query},
\ref{tab:imple_abla_prior_query}.

\textbf{Training configurations on DataComp-1B.}
As shown in Tab.~\ref{tab:imple_abla_linear_128},
\ref{tab:imple_abla_attn_128},
\ref{tab:imple_abla_attn_1024}.

\textbf{Hybrid architecture.}
As shown in Tab.~\ref{tab:imple_abla_cnn},
\ref{tab:imple_abla_trans},
\ref{tab:imple_abla_hybrid}.

\textbf{BitDance-T.}
As detailed in Sec.~\ref{sec:method} and \ref{sec:experiments}, 
BitDance-T is an ultra-lightweight model with merely 8.6M parameters. 
Consequently, 
the additional computational overhead incurred during the training phase is negligible.

\textbf{Inference.}
During the inference phase, 
the semantic teacher and generative prior model are not required. 
Consequently, 
UniWeTok relies solely on a single encoder and decoder.

\begin{table}[!htbp]
\centering
\begin{minipage}[tl]{0.28\linewidth}
    \raggedright
    \caption{ \textbf{\textit{w/o.} PPD \textit{w/o.} GAP setting.}
    }
    \belowrulesep=0pt\aboverulesep=0pt
        \resizebox{\textwidth}{!}{
        \begin{tabular}{l|c}
        config & \textit{w/o.} PPD \textit{w/o.} GAP \\
        \Xhline{1.0pt}
        training data & IN-1K training set \\
        image size & [256, 256] \\
        data augmentation & random crop \\
        downsample & $32 \times 32$ \\
        ema & True \\
        $g$ (group number) & 16 \\
        $d'$ (group channel) & 8 \\
        optimizer & Adam \\ 
        optimizer momentum & $\beta_1, \beta_2{=}0.5, 0.9$  \\
        weight decay & 0 \\
        learning rate schedule & consistent \\
        learning rate & 1e-4 \\
        warmup steps & 0 \\
        cos decay end ratio & 1 \\
        total steps &  250250 \\
        channel\_mult & [1,1,2,2,4,8] \\
        channel & 128 \\
        num\_res\_blocks & 2 \\
        bottleneck channel double & True \\
        num\_attn\_blocks & 16 \\
        generative decoder & True \\
        semantic teacher & -- \\
        SigLu activation & False \\
        pre distillation & False \\
        post distillation & False \\
        distill head & linear \\
        prior model & -- \\
        query token & False \\
        global batchsize & 128 \\
        \end{tabular}
    }
    \label{tab:imple_abla_wetok_loss_config}
\end{minipage}
\hfill
\begin{minipage}[tl]{0.32\linewidth}
    \raggedright
    \caption{ \textbf{\textit{w/.} PPD \textit{w/o.} GAP setting.}
    }
    \belowrulesep=0pt\aboverulesep=0pt
        \resizebox{\textwidth}{!}{
        \begin{tabular}{l|c}
        config & \textit{w/.} PPD \textit{w/o.} GAP \quad \\
        \Xhline{1.0pt}
        training data & IN-1K training set \\
        image size & [256, 256] \\
        data augmentation & random crop \\
        downsample & $32 \times 32$ \\
        ema & True \\
        $g$ (group number) & 16 \\
        $d'$ (group channel) & 8 \\
        optimizer & Adam \\ 
        optimizer momentum & $\beta_1, \beta_2{=}0.5, 0.9$  \\
        weight decay & 0 \\
        learning rate schedule & consistent \\
        learning rate & 1e-4 \\
        warmup steps & 0 \\
        cos decay end ratio & 1 \\
        total steps &  250250 \\
        channel\_mult & [1,1,2,2,4,8] \\
        channel & 128 \\
        num\_res\_blocks & 2 \\
        bottleneck channel double & True \\
        num\_attn\_blocks & 16 \\
        generative decoder & True \\
        semantic teacher & ViT-SO400M-16-SigLIP2-384 \\
        SigLu activation & True \\
        pre distillation & True \\
        post distillation & True \\
        distill head & linear \\
        prior model & -- \\
        query token & False \\
        global batchsize & 128 \\
        \end{tabular}
    }
    \label{tab:imple_abla_dis_loss_config}
\end{minipage}
\hfill
\begin{minipage}[tl]{0.32\linewidth}
    \raggedright
    \caption{ \textbf{\textit{w/.} PPD \textit{w/.} GAP setting.}
    }
    \belowrulesep=0pt\aboverulesep=0pt
        \resizebox{\textwidth}{!}{
        \begin{tabular}{l|c}
        config & \quad \textit{w/.} PPD \textit{w/.} GAP \quad \\
        \Xhline{1.0pt}
        training data & IN-1K training set \\
        image size & [256, 256] \\
        data augmentation & random crop \\
        downsample & $32 \times 32$ \\
        ema & True \\
        $g$ (group number) & 16 \\
        $d'$ (group channel) & 8 \\
        optimizer & Adam \\ 
        optimizer momentum & $\beta_1, \beta_2{=}0.5, 0.9$  \\
        weight decay & 0 \\
        learning rate schedule & consistent \\
        learning rate & 1e-4 \\
        warmup steps & 0 \\
        cos decay end ratio & 1 \\
        total steps &  250250 \\
        channel\_mult & [1,1,2,2,4,8] \\
        channel & 128 \\
        num\_res\_blocks & 2 \\
        bottleneck channel double & True \\
        num\_attn\_blocks & 16 \\
        generative decoder & True \\
        semantic teacher & ViT-SO400M-16-SigLIP2-384 \\
        SigLu activation & True \\
        pre distillation & True \\
        post distillation & True \\
        distill head & linear \\
        prior model & BitDance-T \\
        query token & True \\
        global batchsize & 128 \\
        \end{tabular}
    }
    \label{tab:imple_abla_uniwetok_loss_config}
\end{minipage}
\end{table}
\begin{table}[!htbp]
\centering
\begin{minipage}[tl]{0.32\linewidth}
    \raggedright
    \caption{\textbf{Pre distillation setting.}
    }
    \belowrulesep=0pt\aboverulesep=0pt
        \resizebox{\textwidth}{!}{
        \begin{tabular}{l|c}
        config & pre distillation \\
        \Xhline{1.0pt}
        training data & IN-1K training set \\
        image size & [256, 256] \\
        data augmentation & random crop \\
        downsample & $32 \times 32$ \\
        ema & True \\
        $g$ (group number) & 16 \\
        $d'$ (group channel) & 8 \\
        optimizer & Adam \\ 
        optimizer momentum & $\beta_1, \beta_2{=}0.5, 0.9$  \\
        weight decay & 0 \\
        learning rate schedule & consistent \\
        learning rate & 1e-4 \\
        warmup steps & 0 \\
        cos decay end ratio & 1 \\
        total steps &  250250 \\
        channel\_mult & [1,1,2,2,4,8] \\
        channel & 128 \\
        num\_res\_blocks & 2 \\
        bottleneck channel double & True \\
        num\_attn\_blocks & 16 \\
        generative decoder & True \\
        semantic teacher & ViT-SO400M-16-SigLIP2-384 \\
        SigLu activation & False \\
        pre distillation & True \\
        post distillation & False \\
        distill head & linear \\
        prior model & -- \\
        query token & False \\
        global batchsize & 128 \\
        \end{tabular}
    }
    \label{tab:imple_abla_nosiglu_pre}
\end{minipage}
\hfill
\begin{minipage}[tl]{0.32\linewidth}
    \raggedright
    \caption{\textbf{Post distillation setting.}
    }
    \belowrulesep=0pt\aboverulesep=0pt
        \resizebox{\textwidth}{!}{
        \begin{tabular}{l|c}
        config & post distillation \\
        \Xhline{1.0pt}
        training data & IN-1K training set \\
        image size & [256, 256] \\
        data augmentation & random crop \\
        downsample & $32 \times 32$ \\
        ema & True \\
        $g$ (group number) & 16 \\
        $d'$ (group channel) & 8 \\
        optimizer & Adam \\ 
        optimizer momentum & $\beta_1, \beta_2{=}0.5, 0.9$  \\
        weight decay & 0 \\
        learning rate schedule & consistent \\
        learning rate & 1e-4 \\
        warmup steps & 0 \\
        cos decay end ratio & 1 \\
        total steps &  250250 \\
        channel\_mult & [1,1,2,2,4,8] \\
        channel & 128 \\
        num\_res\_blocks & 2 \\
        bottleneck channel double & True \\
        num\_attn\_blocks & 16 \\
        generative decoder & True \\
        semantic teacher & ViT-SO400M-16-SigLIP2-384 \\
        SigLu activation & False \\
        pre distillation & False \\
        post distillation & True \\
        distill head & linear \\
        prior model & -- \\
        query token & False \\
        global batchsize & 128 \\
        \end{tabular}
    }
    \label{tab:imple_abla_nosiglu_post}
\end{minipage}
\hfill
\begin{minipage}[tl]{0.32\linewidth}
    \raggedright
    \caption{\textbf{Post distillation with SigLu activation setting.}
    }
    \belowrulesep=0pt\aboverulesep=0pt
        \resizebox{\textwidth}{!}{
        \begin{tabular}{l|c}
        config & post distillation with SigLu \\
        \Xhline{1.0pt}
        training data & IN-1K training set \\
        image size & [256, 256] \\
        data augmentation & random crop \\
        downsample & $32 \times 32$ \\
        ema & True \\
        $g$ (group number) & 16 \\
        $d'$ (group channel) & 8 \\
        optimizer & Adam \\ 
        optimizer momentum & $\beta_1, \beta_2{=}0.5, 0.9$  \\
        weight decay & 0 \\
        learning rate schedule & consistent \\
        learning rate & 1e-4 \\
        warmup steps & 0 \\
        cos decay end ratio & 1 \\
        total steps &  250250 \\
        channel\_mult & [1,1,2,2,4,8] \\
        channel & 128 \\
        num\_res\_blocks & 2 \\
        bottleneck channel double & True \\
        num\_attn\_blocks & 16 \\
        generative decoder & True \\
        semantic teacher & ViT-SO400M-16-SigLIP2-384 \\
        SigLu activation & True \\
        pre distillation & False \\
        post distillation & True \\
        distill head & linear \\
        prior model & -- \\
        query token & False \\
        global batchsize & 128 \\
        \end{tabular}
    }
    \label{tab:imple_abla_siglu_post}
\end{minipage}
\end{table}
\begin{table}[!htbp]
\centering
\begin{minipage}[tl]{0.32\linewidth}
    \raggedright
    \caption{\textbf{Pre distillation setting.}
    }
    \belowrulesep=0pt\aboverulesep=0pt
        \resizebox{\textwidth}{!}{
        \begin{tabular}{l|c}
        config & pre distillation \\
        \Xhline{1.0pt}
        training data & IN-1K training set \\
        image size & [256, 256] \\
        data augmentation & random crop \\
        downsample & $32 \times 32$ \\
        ema & True \\
        $g$ (group number) & 16 \\
        $d'$ (group channel) & 8 \\
        optimizer & Adam \\ 
        optimizer momentum & $\beta_1, \beta_2{=}0.5, 0.9$  \\
        weight decay & 0 \\
        learning rate schedule & consistent \\
        learning rate & 1e-4 \\
        warmup steps & 0 \\
        cos decay end ratio & 1 \\
        total steps &  250250 \\
        channel\_mult & [1,1,2,2,4,8] \\
        channel & 128 \\
        num\_res\_blocks & 2 \\
        bottleneck channel double & True \\
        num\_attn\_blocks & 16 \\
        generative decoder & True \\
        semantic teacher & ViT-SO400M-16-SigLIP2-384 \\
        SigLu activation & True \\
        pre distillation & True \\
        post distillation & False \\
        distill head & linear \\
        prior model & -- \\
        query token & False \\
        global batchsize & 128 \\
        \end{tabular}
    }
    \label{tab:imple_abla_pre_dis}
\end{minipage}
\hfill
\begin{minipage}[tl]{0.32\linewidth}
    \raggedright
    \caption{\textbf{Post distillation setting.}
    }
    \belowrulesep=0pt\aboverulesep=0pt
        \resizebox{\textwidth}{!}{
        \begin{tabular}{l|c}
        config & post distillation \\
        \Xhline{1.0pt}
        training data & IN-1K training set \\
        image size & [256, 256] \\
        data augmentation & random crop \\
        downsample & $32 \times 32$ \\
        ema & True \\
        $g$ (group number) & 16 \\
        $d'$ (group channel) & 8 \\
        optimizer & Adam \\ 
        optimizer momentum & $\beta_1, \beta_2{=}0.5, 0.9$  \\
        weight decay & 0 \\
        learning rate schedule & consistent \\
        learning rate & 1e-4 \\
        warmup steps & 0 \\
        cos decay end ratio & 1 \\
        total steps &  250250 \\
        channel\_mult & [1,1,2,2,4,8] \\
        channel & 128 \\
        num\_res\_blocks & 2 \\
        bottleneck channel double & True \\
        num\_attn\_blocks & 16 \\
        generative decoder & True \\
        semantic teacher & ViT-SO400M-16-SigLIP2-384 \\
        SigLu activation & True \\
        pre distillation & False \\
        post distillation & True \\
        distill head & linear \\
        prior model & -- \\
        query token & False \\
        global batchsize & 128 \\
        \end{tabular}
    }
    \label{tab:imple_abla_post_dis}
\end{minipage}
\hfill
\begin{minipage}[tl]{0.32\linewidth}
    \raggedright
    \caption{\textbf{Pre-Post distillation setting.}
    }
    \belowrulesep=0pt\aboverulesep=0pt
        \resizebox{\textwidth}{!}{
        \begin{tabular}{l|c}
        config & pre-post distillation \\
        \Xhline{1.0pt}
        training data & IN-1K training set \\
        image size & [256, 256] \\
        data augmentation & random crop \\
        downsample & $32 \times 32$ \\
        ema & True \\
        $g$ (group number) & 16 \\
        $d'$ (group channel) & 8 \\
        optimizer & Adam \\ 
        optimizer momentum & $\beta_1, \beta_2{=}0.5, 0.9$  \\
        weight decay & 0 \\
        learning rate schedule & consistent \\
        learning rate & 1e-4 \\
        warmup steps & 0 \\
        cos decay end ratio & 1 \\
        total steps &  250250 \\
        channel\_mult & [1,1,2,2,4,8] \\
        channel & 128 \\
        num\_res\_blocks & 2 \\
        bottleneck channel double & True \\
        num\_attn\_blocks & 16 \\
        generative decoder & True \\
        semantic teacher & ViT-SO400M-16-SigLIP2-384 \\
        SigLu activation & True \\
        pre distillation & True \\
        post distillation & True \\
        distill head & linear \\
        prior model & -- \\
        query token & False \\
        global batchsize & 128 \\
        \end{tabular}
    }
    \label{tab:imple_abla_pre_post_dis}
\end{minipage}
\end{table}
\begin{table}[!htbp]
\centering
\begin{minipage}[tl]{0.48\linewidth}
    \raggedright
    \caption{\textbf{Single bottleneck channel setting.}
    }
    \belowrulesep=0pt\aboverulesep=0pt
        \resizebox{\textwidth}{!}{
        \begin{tabular}{l|c}
        config & single bottleneck channel \\
        \Xhline{1.0pt}
        training data & IN-1K training set \\
        image size & [256, 256] \\
        data augmentation & random crop \\
        downsample & $32 \times 32$ \\
        ema & True \\
        $g$ (group number) & 16 \\
        $d'$ (group channel) & 8 \\
        optimizer & Adam \\ 
        optimizer momentum & $\beta_1, \beta_2{=}0.5, 0.9$  \\
        weight decay & 0 \\
        learning rate schedule & consistent \\
        learning rate & 1e-4 \\
        warmup steps & 0 \\
        cos decay end ratio & 1 \\
        total steps &  250250 \\
        channel\_mult & [1,1,2,2,4,8] \\
        channel & 128 \\
        num\_res\_blocks & 2 \\
        bottleneck channel double & False \\
        num\_attn\_blocks & 16 \\
        generative decoder & True \\
        semantic teacher & ViT-SO400M-16-SigLIP2-384 \\
        SigLu activation & True \\
        pre distillation & True \\
        post distillation & True \\
        distill head & linear \\
        prior model & -- \\
        query token & False \\
        global batchsize & 128 \\
        \end{tabular}
    }
    \label{tab:imple_abla_single_ch}
\end{minipage}
\hfill
\begin{minipage}[tl]{0.48\linewidth}
    \raggedright
    \caption{\textbf{Double bottleneck channel setting.}
    }
    \belowrulesep=0pt\aboverulesep=0pt
        \resizebox{\textwidth}{!}{
        \begin{tabular}{l|c}
        config & double bottleneck channel \\
        \Xhline{1.0pt}
        training data & IN-1K training set \\
        image size & [256, 256] \\
        data augmentation & random crop \\
        downsample & $32 \times 32$ \\
        ema & True \\
        $g$ (group number) & 16 \\
        $d'$ (group channel) & 8 \\
        optimizer & Adam \\ 
        optimizer momentum & $\beta_1, \beta_2{=}0.5, 0.9$  \\
        weight decay & 0 \\
        learning rate schedule & consistent \\
        learning rate & 1e-4 \\
        warmup steps & 0 \\
        cos decay end ratio & 1 \\
        total steps &  250250 \\
        channel\_mult & [1,1,2,2,4,8] \\
        channel & 128 \\
        num\_res\_blocks & 2 \\
        bottleneck channel double & True \\
        num\_attn\_blocks & 16 \\
        generative decoder & True \\
        semantic teacher & ViT-SO400M-16-SigLIP2-384 \\
        SigLu activation & True \\
        pre distillation & True \\
        post distillation & True \\
        distill head & linear \\
        prior model & -- \\
        query token & False \\
        global batchsize & 128 \\
        \end{tabular}
    }
    \label{tab:imple_abla_double_ch}
\end{minipage}
\end{table}
\begin{table}[!htbp]
\centering
\begin{minipage}[tl]{0.32\linewidth}
    \raggedright
    \caption{ \textbf{\textit{w/o.} GAP \textit{w/o.} Query setting.}
    }
    \belowrulesep=0pt\aboverulesep=0pt
        \resizebox{\textwidth}{!}{
        \begin{tabular}{l|c}
        config & \textit{w/o.} GAP \textit{w/o.} Query \\
        \Xhline{1.0pt}
        training data & IN-1K training set \\
        image size & [256, 256] \\
        data augmentation & random crop \\
        downsample & $32 \times 32$ \\
        ema & True \\
        $g$ (group number) & 16 \\
        $d'$ (group channel) & 8 \\
        optimizer & Adam \\ 
        optimizer momentum & $\beta_1, \beta_2{=}0.5, 0.9$  \\
        weight decay & 0 \\
        learning rate schedule & consistent \\
        learning rate & 1e-4 \\
        warmup steps & 0 \\
        cos decay end ratio & 1 \\
        total steps &  250250 \\
        channel\_mult & [1,1,2,2,4,8] \\
        channel & 128 \\
        num\_res\_blocks & 2 \\
        bottleneck channel double & True \\
        num\_attn\_blocks & 16 \\
        generative decoder & True \\
        semantic teacher & ViT-SO400M-16-SigLIP2-384 \\
        SigLu activation & True \\
        pre distillation & True \\
        post distillation & True \\
        distill head & linear \\
        prior model & -- \\
        query token & False \\
        global batchsize & 128 \\
        \end{tabular}
    }
    \label{tab:imple_abla_no_prior_no_query}
\end{minipage}
\hfill
\begin{minipage}[tl]{0.32\linewidth}
    \raggedright
    \caption{ \textbf{\textit{w/.} GAP \textit{w/o.} Query setting.}
    }
    \belowrulesep=0pt\aboverulesep=0pt
        \resizebox{\textwidth}{!}{
        \begin{tabular}{l|c}
        config & \textit{w/.} GAP \textit{w/o.} Query \\
        \Xhline{1.0pt}
        training data & IN-1K training set \\
        image size & [256, 256] \\
        data augmentation & random crop \\
        downsample & $32 \times 32$ \\
        ema & True \\
        $g$ (group number) & 16 \\
        $d'$ (group channel) & 8 \\
        optimizer & Adam \\ 
        optimizer momentum & $\beta_1, \beta_2{=}0.5, 0.9$  \\
        weight decay & 0 \\
        learning rate schedule & consistent \\
        learning rate & 1e-4 \\
        warmup steps & 0 \\
        cos decay end ratio & 1 \\
        total steps &  250250 \\
        channel\_mult & [1,1,2,2,4,8] \\
        channel & 128 \\
        num\_res\_blocks & 2 \\
        bottleneck channel double & True \\
        num\_attn\_blocks & 16 \\
        generative decoder & True \\
        semantic teacher & ViT-SO400M-16-SigLIP2-384 \\
        SigLu activation & True \\
        pre distillation & True \\
        post distillation & True \\
        distill head & linear \\
        prior model & BitDance-T \\
        query token & False \\
        global batchsize & 128 \\
        \end{tabular}
    }
    \label{tab:imple_abla_prior_no_query}
\end{minipage}
\hfill
\begin{minipage}[tl]{0.32\linewidth}
    \raggedright
    \caption{ \textbf{\textit{w/.} GAP \textit{w/.} Query setting.}
    }
    \belowrulesep=0pt\aboverulesep=0pt
        \resizebox{\textwidth}{!}{
        \begin{tabular}{l|c}
        config & \textit{w/.} GAP \textit{w/.} Query \\
        \Xhline{1.0pt}
        training data & IN-1K training set \\
        image size & [256, 256] \\
        data augmentation & random crop \\
        downsample & $32 \times 32$ \\
        ema & True \\
        $g$ (group number) & 16 \\
        $d'$ (group channel) & 8 \\
        optimizer & Adam \\ 
        optimizer momentum & $\beta_1, \beta_2{=}0.5, 0.9$  \\
        weight decay & 0 \\
        learning rate schedule & consistent \\
        learning rate & 1e-4 \\
        warmup steps & 0 \\
        cos decay end ratio & 1 \\
        total steps &  250250 \\
        channel\_mult & [1,1,2,2,4,8] \\
        channel & 128 \\
        num\_res\_blocks & 2 \\
        bottleneck channel double & True \\
        num\_attn\_blocks & 16 \\
        generative decoder & True \\
        semantic teacher & ViT-SO400M-16-SigLIP2-384 \\
        SigLu activation & True \\
        pre distillation & True \\
        post distillation & True \\
        prior model & BitDance-T \\
        query token & True \\
        global batchsize & 128 \\
        \end{tabular}
    }
    \label{tab:imple_abla_prior_query}
\end{minipage}
\end{table}
\begin{table}[!htbp]
\centering
\begin{minipage}[tl]{0.32\linewidth}
    \raggedright
    \caption{\textbf{Linear head with batchsize 128 setting.}
    }
    \belowrulesep=0pt\aboverulesep=0pt
        \resizebox{\textwidth}{!}{
        \begin{tabular}{l|c}
        config & linear head with bs128 \\
        \Xhline{1.0pt}
        training data & DataComp-1B \\
        image size & [256, 256] \\
        data augmentation & random crop \\
        downsample & $32 \times 32$ \\
        ema & True \\
        $g$ (group number) & 16 \\
        $d'$ (group channel) & 8 \\
        optimizer & Adam \\ 
        optimizer momentum & $\beta_1, \beta_2{=}0.5, 0.9$  \\
        weight decay & 0 \\
        learning rate schedule & consistent \\
        learning rate & 1e-4 \\
        warmup steps & 0 \\
        cos decay end ratio & 1 \\
        total steps &  250250 \\
        channel\_mult & [1,1,2,2,4,8] \\
        channel & 128 \\
        num\_res\_blocks & 2 \\
        bottleneck channel double & True \\
        num\_attn\_blocks & 0 \\
        generative decoder & True \\
        semantic teacher & ViT-SO400M-16-SigLIP2-384 \\
        SigLu activation & True \\
        pre distillation & True \\
        post distillation & True \\
        distill head & linear \\
        prior model & BitDance-T \\
        query token & True \\
        global batchsize & 128 \\
        \end{tabular}
    }
    \label{tab:imple_abla_linear_128}
\end{minipage}
\hfill
\begin{minipage}[tl]{0.32\linewidth}
    \raggedright
    \caption{\textbf{Attention head with batchsize 128 setting.}
    }
    \belowrulesep=0pt\aboverulesep=0pt
        \resizebox{\textwidth}{!}{
        \begin{tabular}{l|c}
        config & attention head with bs1024 \\
        \Xhline{1.0pt}
        training data & DataComp-1B \\
        image size & [256, 256] \\
        data augmentation & random crop \\
        downsample & $32 \times 32$ \\
        ema & True \\
        $g$ (group number) & 16 \\
        $d'$ (group channel) & 8 \\
        optimizer & Adam \\ 
        optimizer momentum & $\beta_1, \beta_2{=}0.5, 0.9$  \\
        weight decay & 0 \\
        learning rate schedule & consistent \\
        learning rate & 1e-4 \\
        warmup steps & 0 \\
        cos decay end ratio & 1 \\
        total steps &  250250 \\
        channel\_mult & [1,1,2,2,4,8] \\
        channel & 128 \\
        num\_res\_blocks & 2 \\
        bottleneck channel double & True \\
        num\_attn\_blocks & 0 \\
        generative decoder & True \\
        semantic teacher & ViT-SO400M-16-SigLIP2-384 \\
        SigLu activation & True \\
        pre distillation & True \\
        post distillation & True \\
        distill head & attention \\
        prior model & BitDance-T \\
        query token & True \\
        global batchsize & 128 \\
        \end{tabular}
    }
    \label{tab:imple_abla_attn_128}
\end{minipage}
\hfill
\begin{minipage}[tl]{0.32\linewidth}
    \raggedright
    \caption{\textbf{Attention head with batchsize 1024 setting.}
    }
    \belowrulesep=0pt\aboverulesep=0pt
        \resizebox{\textwidth}{!}{
        \begin{tabular}{l|c}
        config & attention head with bs1024 \\
        \Xhline{1.0pt}
        training data & DataComp-1B \\
        image size & [256, 256] \\
        data augmentation & random crop \\
        downsample & $32 \times 32$ \\
        ema & True \\
        $g$ (group number) & 16 \\
        $d'$ (group channel) & 8 \\
        optimizer & Adam \\ 
        optimizer momentum & $\beta_1, \beta_2{=}0.5, 0.9$  \\
        weight decay & 0 \\
        learning rate schedule & consistent \\
        learning rate & 1e-4 \\
        warmup steps & 0 \\
        cos decay end ratio & 1 \\
        total steps &  250250 \\
        channel\_mult & [1,1,2,2,4,8] \\
        channel & 128 \\
        num\_res\_blocks & 2 \\
        bottleneck channel double & True \\
        num\_attn\_blocks & 0 \\
        generative decoder & True \\
        semantic teacher & ViT-SO400M-16-SigLIP2-384 \\
        SigLu activation & True \\
        pre distillation & True \\
        post distillation & True \\
        distill head & attention \\
        prior model & BitDance-T \\
        query token & True \\
        global batchsize & 1024 \\
        \end{tabular}
    }
    \label{tab:imple_abla_attn_1024}
\end{minipage}
\end{table}
\begin{table}[!htbp]
\centering
\begin{minipage}[tl]{0.32\linewidth}
    \raggedright
    \caption{\textbf{CNN architecture setting.}
    }
    \belowrulesep=0pt\aboverulesep=0pt
        \resizebox{\textwidth}{!}{
        \begin{tabular}{l|c}
        config & CNN architecture \\
        \Xhline{1.0pt}
        training data & DataComp-1B \\
        image size & [256, 256] \\
        data augmentation & random crop \\
        downsample & $32 \times 32$ \\
        ema & True \\
        $g$ (group number) & 16 \\
        $d'$ (group channel) & 8 \\
        optimizer & Adam \\ 
        optimizer momentum & $\beta_1, \beta_2{=}0.5, 0.9$  \\
        weight decay & 0 \\
        learning rate schedule & consistent \\
        learning rate & 1e-4 \\
        warmup steps & 0 \\
        cos decay end ratio & 1 \\
        total steps &  250250 \\
        channel\_mult & [1,1,2,2,4,8] \\
        channel & 128 \\
        num\_res\_blocks & 2 \\
        bottleneck channel double & True \\
        num\_attn\_blocks & 0 \\
        generative decoder & True \\
        semantic teacher & ViT-SO400M-16-SigLIP2-384 \\
        SigLu activation & True \\
        pre distillation & True \\
        post distillation & True \\
        distill head & attention \\
        prior model & BitDance-T \\
        query token & True \\
        global batchsize & 1024 \\
        \end{tabular}
    }
    \label{tab:imple_abla_cnn}
\end{minipage}
\hfill
\begin{minipage}[tl]{0.32\linewidth}
    \raggedright
    \caption{\textbf{Transformer architecture setting.}
    }
    \belowrulesep=0pt\aboverulesep=0pt
        \resizebox{\textwidth}{!}{
        \begin{tabular}{l|c}
        config & transformer architecture \\
        \Xhline{1.0pt}
        training data & DataComp-1B \\
        image size & [256, 256] \\
        data augmentation & random crop \\
        downsample & $32 \times 32$ \\
        ema & True \\
        $g$ (group number) & 16 \\
        $d'$ (group channel) & 8 \\
        optimizer & Adam \\ 
        optimizer momentum & $\beta_1, \beta_2{=}0.5, 0.9$  \\
        weight decay & 0 \\
        learning rate schedule & consistent \\
        learning rate & 1e-4 \\
        warmup steps & 0 \\
        cos decay end ratio & 1 \\
        total steps &  250250 \\
        channel\_mult & [1,1,2,2,4,8] \\
        channel & 128 \\
        num\_res\_blocks & 0 \\
        bottleneck channel double & True \\
        num\_attn\_blocks & 16 \\
        generative decoder & True \\
        semantic teacher & ViT-SO400M-16-SigLIP2-384 \\
        SigLu activation & True \\
        pre distillation & True \\
        post distillation & True \\
        distill head & attention \\
        prior model & BitDance-T \\
        query token & True \\
        global batchsize & 1024 \\
        \end{tabular}
    }
    \label{tab:imple_abla_trans}
\end{minipage}
\hfill
\begin{minipage}[tl]{0.32\linewidth}
    \raggedright
    \caption{\textbf{Hybrid architecture setting.}
    }
    \belowrulesep=0pt\aboverulesep=0pt
        \resizebox{\textwidth}{!}{
        \begin{tabular}{l|c}
        config & hybrid architecture \\
        \Xhline{1.0pt}
        training data & DataComp-1B \\
        image size & [256, 256] \\
        data augmentation & random crop \\
        downsample & $32 \times 32$ \\
        ema & True \\
        $g$ (group number) & 16 \\
        $d'$ (group channel) & 8 \\
        optimizer & Adam \\ 
        optimizer momentum & $\beta_1, \beta_2{=}0.5, 0.9$  \\
        weight decay & 0 \\
        learning rate schedule & consistent \\
        learning rate & 1e-4 \\
        warmup steps & 0 \\
        cos decay end ratio & 1 \\
        total steps &  250250 \\
        channel\_mult & [1,1,2,2,4,8] \\
        channel & 128 \\
        num\_res\_blocks & 2 \\
        bottleneck channel double & True \\
        num\_attn\_blocks & 16 \\
        generative decoder & True \\
        semantic teacher & ViT-SO400M-16-SigLIP2-384 \\
        SigLu activation & True \\
        pre distillation & True \\
        post distillation & True \\
        distill head & attention \\
        prior model & BitDance-T \\
        query token & True \\
        global batchsize & 1024 \\
        \end{tabular}
    }
    \label{tab:imple_abla_hybrid}
\end{minipage}
\end{table}

\section{More Visualization Results}
\label{app:vis_result}

\textbf{Multimodal Understanding}.
More visualization results of multimodal understanding by UniWeTok-Chat could be found in Fig.~\ref{}.

\begin{figure*}[!ht]
    \centering
    \includegraphics[width=0.85\linewidth]{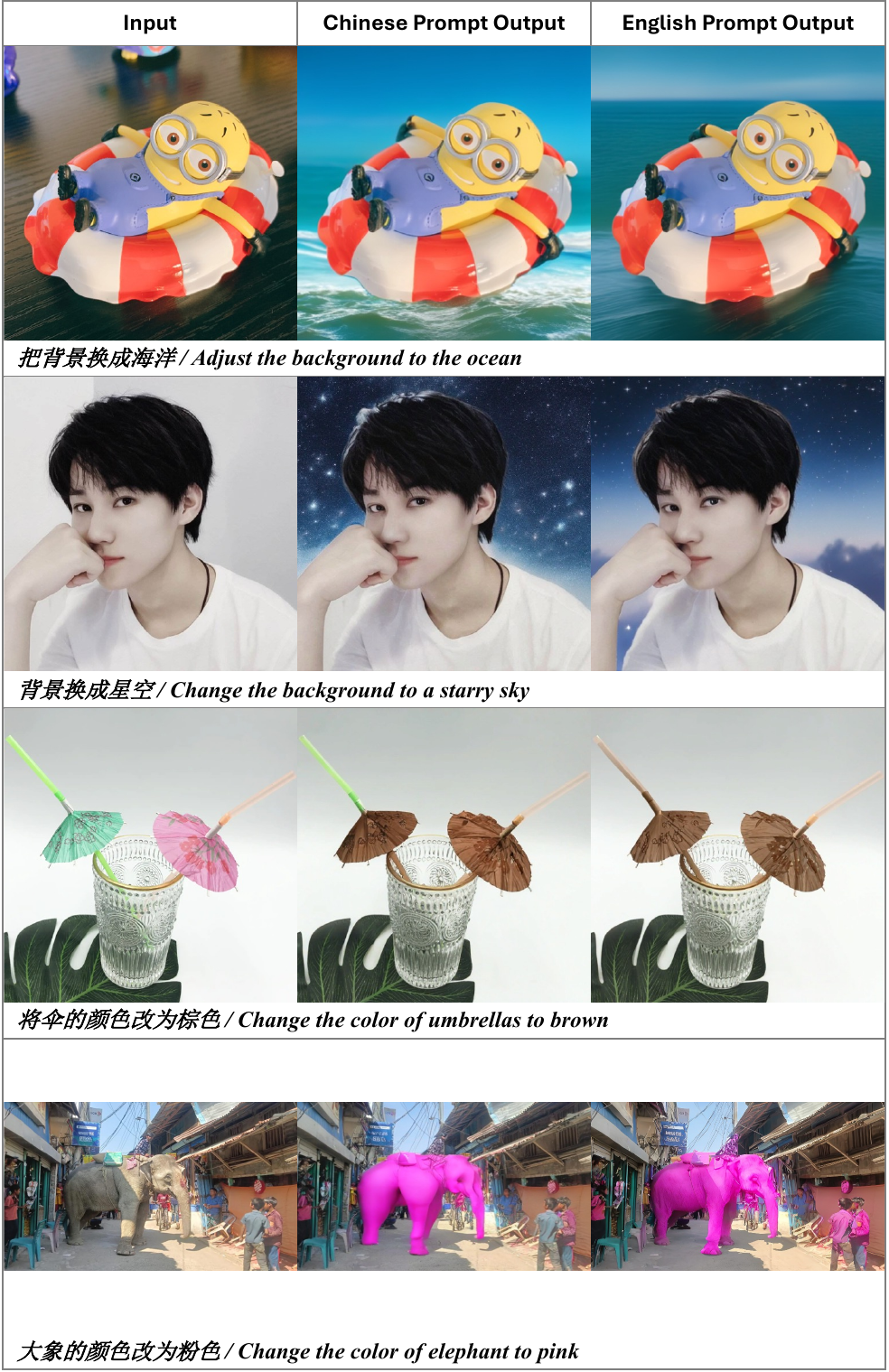}
    \caption{\textbf{Visualization results of image editing by UniWeTok-Edit part 1.}
    }
    \label{fig:vis_edit1}
\end{figure*}

\begin{figure*}[!ht]
    \centering
    \includegraphics[width=0.85\linewidth]{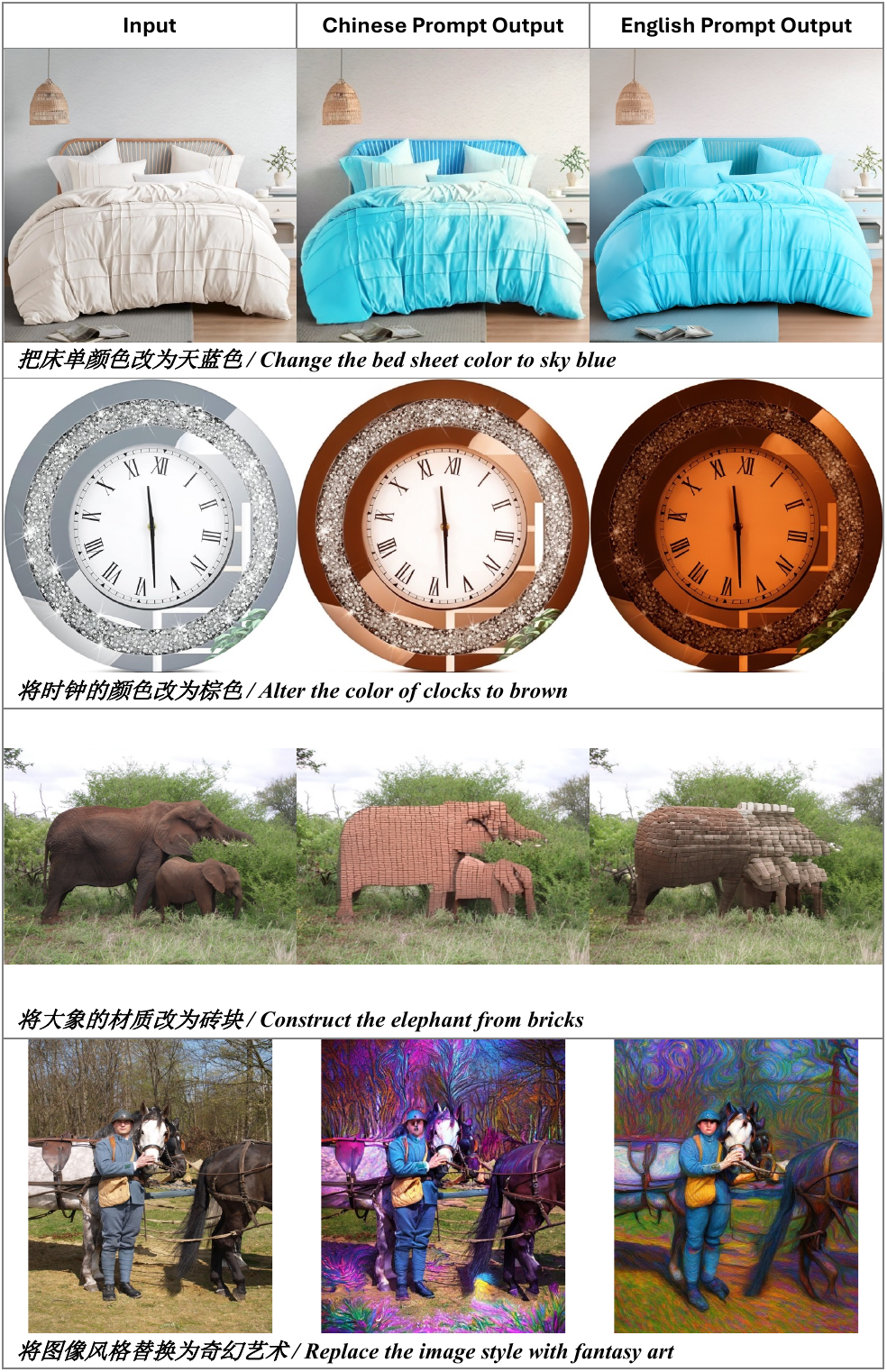}
    \caption{\textbf{Visualization results of image editing by UniWeTok-Edit part 2.}
    }
    \label{fig:vis_edit2}
\end{figure*}

\begin{figure*}[!ht]
    \centering
    \includegraphics[width=0.85\linewidth]{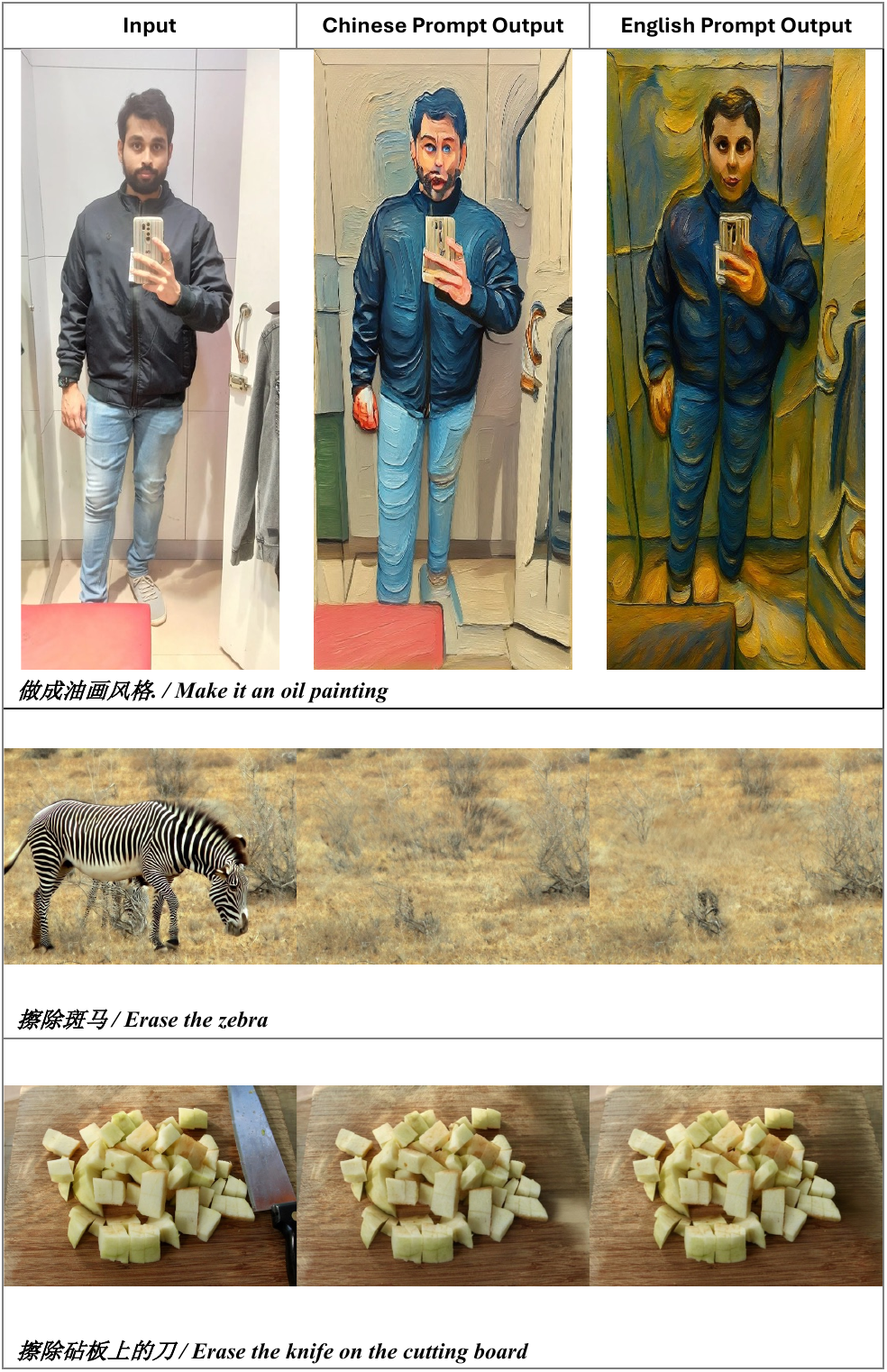}
    \caption{\textbf{Visualization results of image editing by UniWeTok-Edit part 3.}
    }
    \label{fig:vis_edit3}
\end{figure*}

\textbf{Image Editing}.
More visualization results of image editing by UniWeTok-Edit could be found in Fig.~\ref{fig:vis_edit1},\ref{fig:vis_edit2},\ref{fig:vis_edit3}.

\begin{figure*}[!ht]
    \centering
    \includegraphics[width=0.98\linewidth]{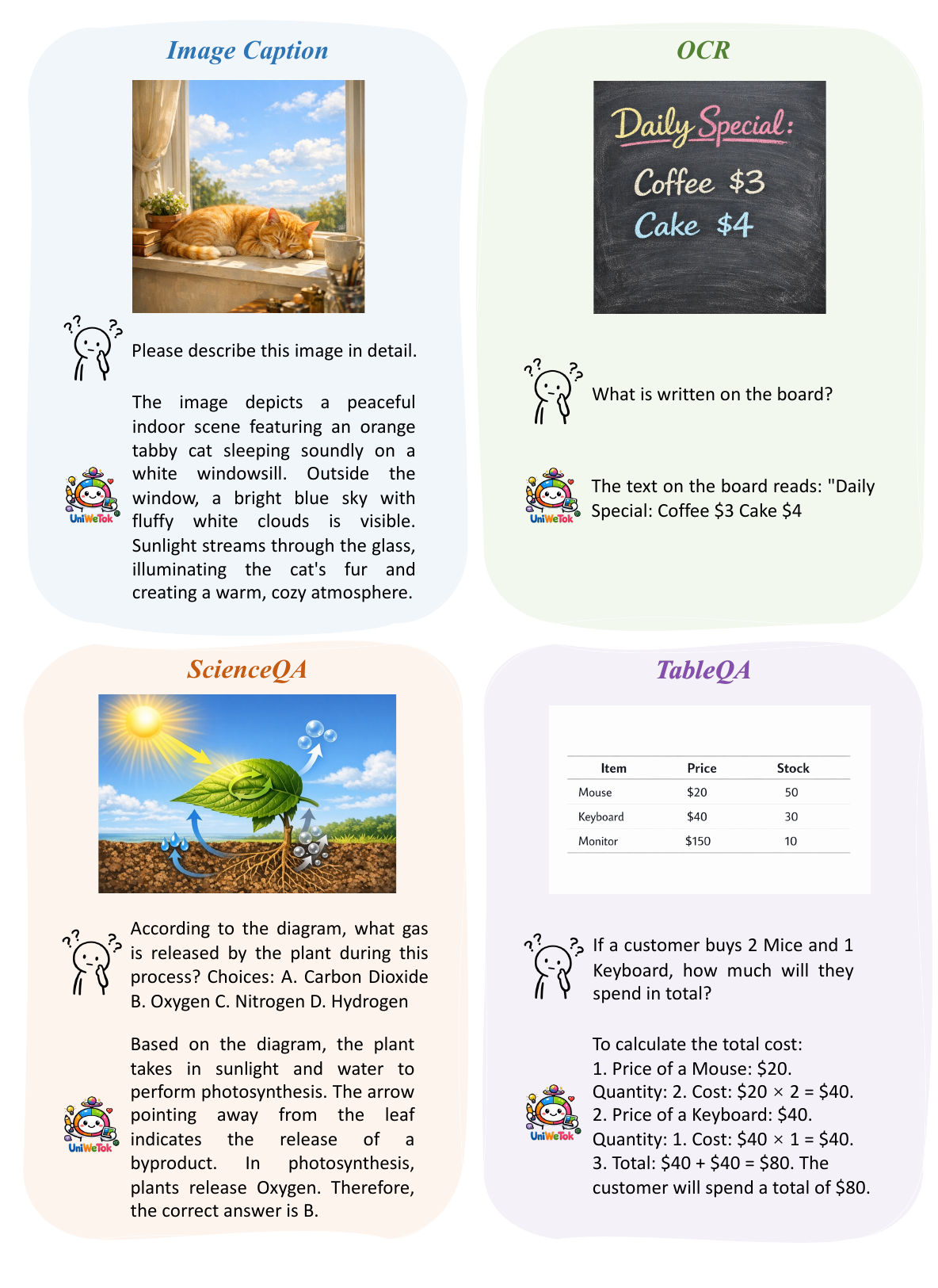}
    \caption{\textbf{Visualization results of multimodal understanding by UniWeTok-Chat.}
    }
    \label{fig:vis_und}
\end{figure*}

\end{document}